\documentclass{article}

\usepackage{PRIMEarxiv}

\usepackage[utf8]{inputenc} 
\usepackage[T1]{fontenc}    
\usepackage{hyperref}       
\usepackage{url}            
\usepackage{booktabs}       
\usepackage{amsfonts}       
\usepackage{nicefrac}       
\usepackage{microtype}      
\usepackage{lipsum}
\usepackage{graphicx}
\graphicspath{{media/}}     
\usepackage{subcaption}
\usepackage{float}
\usepackage{esvect}
\usepackage{breqn}

\usepackage[mathscr]{euscript}

\title{PICS in Pics: Physics Informed Contour Selection for Rapid Image Segmentation 
}

\author{Vikas Dwivedi\\
	Atmospheric Science Research Center\\
	State University of New York, Albany\\
	New York, 12222, USA \\
	\texttt{vdwivedi@albany.edu} \\
	\And
	Balaji Srinivasan \\
	Department of Mechanical Engineering\\
	Indian Institute of Technology, Madras\\
	Chennai, 600036, India \\
	\texttt{sbalaji@iitm.ac.in} \\
	\AND
	Ganapathy Krishnamurthi \\
	Department of Engineering Design \\
	Indian Institute of Technology, Madras\\
	Chennai, 600036, India \\
	\texttt{gankrish@iitm.ac.in} \\
}

\begin{document}
	\maketitle

	\begin{abstract}
		Effective training of deep image segmentation models is challenging due to the need for abundant, high-quality annotations. Generating annotations is laborious and time-consuming for human experts, especially in medical image segmentation. To facilitate image annotation, we introduce Physics Informed Contour Selection (PICS) - an interpretable, physics-informed algorithm for rapid image segmentation without relying on labeled data. PICS draws inspiration from physics-informed neural networks (PINNs) and an active contour model called snake. It is fast and computationally lightweight because it employs cubic splines instead of a deep neural network as a basis function. Its training parameters are physically interpretable because they directly represent control knots of the segmentation curve. Traditional snakes involve minimization of the edge-based loss functionals by deriving the Euler-Lagrange equation followed by its numerical solution. However, PICS directly minimizes the loss functional, bypassing the Euler Lagrange equations. It is the first snake variant to minimize a region-based loss function instead of traditional edge-based loss functions. PICS uniquely models the three-dimensional (3D) segmentation process with an unsteady partial differential equation (PDE), which allows accelerated segmentation via transfer learning. To demonstrate its effectiveness, we apply PICS for 3D segmentation of the left ventricle on a publicly available cardiac dataset. While doing so, we also introduce a new convexity-preserving loss term that encodes the shape information of the left ventricle to enhance PICS's segmentation quality. Overall, PICS presents several novelties in network architecture, transfer learning, and physics-inspired losses for image segmentation, thereby showing promising outcomes and potential for further refinement.
	\end{abstract}


	\section{Introduction}
	
	Deep learning-based computer vision models have achieved remarkable success in various medical imaging tasks. However, their reliance on large amounts of labeled data can limit their utility in situations where data is scarce or unavailable. This limitation has spurred significant recent advancements in the field of physics-informed computer vision (PICV), as discussed in Banerjee et al. \cite{banerjee2023physicsinformed}.
	
	The term "physics-informed" in PICV is largely attributed to the development of the Physics Informed Neural Network (PINN) by Raissi et al. \cite{RAISSI2019}. PINNs have shown promise in addressing forward and inverse problems related to partial differential equations (PDEs) in fields like fluid mechanics \cite{raissi2018hidden, dwivedi2020physics}, material modeling \cite{Liu2019}, heat transfer \cite{cai2021, dwivedi2021distributed}, and more \cite{Karniadakis2021}.
	
	Inspired by the philosophy behind PINNs, PICV integrates physical principles into machine learning frameworks for computer vision. This approach results in faster, more interpretable, and data-efficient computer vision models. In context of medical imaging, some recent examples are as follows. Lopez et al.\cite{ARRATIALOPEZ2023102925} recently introduced WarpPINN, a physics-informed neural network designed for image registration to assess local metrics of heart deformation. They incorporated the near-incompressibility of cardiac tissue by penalizing the Jacobian of the deformation field. Similarly, Vries et al.\cite{DEVRIES2023102971} developed a PINN-based model to estimate CT perfusion parameters from noisy data related to acute ischemic stroke. Herten et al.\cite{VANHERTEN2022102399}  utilized PINNs for tracer-kinetic modeling and parameter inference using myocardial perfusion medical resonance imaging (MRI) data. Buoso and collaborators proposed a parametric PINN for simulating personalized left-ventricular biomechanics, offering the potential to significantly expedite training data synthesis \cite{BUOSO2021102066}. Additionally, Burwinkel et al.\cite{BURWINKEL2022102314} introduced OpticNet, an innovative optical refraction network that utilizes an unsupervised, domain-specific loss function to explicitly incorporate ophthalmological information into the network.

	Our primary focus in this study is image segmentation, which involves identifying and delineating distinct regions or objects within an image. The approaches to image segmentation can be broadly categorized into two extremes: deep learning-based models that rely on substantial labeled training data and traditional models that do not require training data but face challenges related to some theoretical and numerical aspects.

	\begin{figure}[ht!]
		\centering
		\includegraphics[width=0.9\linewidth]{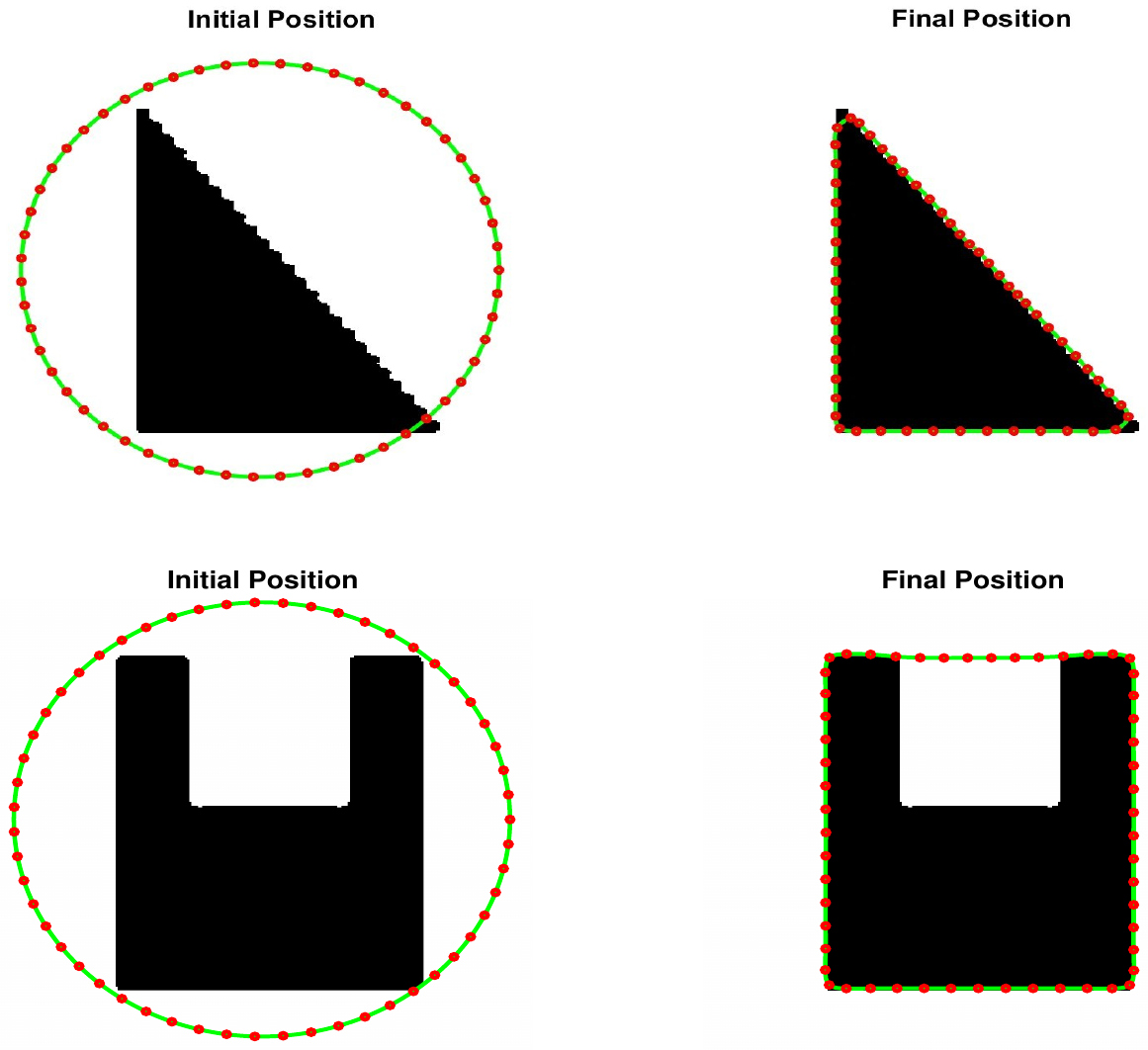}
		\caption{Image segmentation with a naive image gradient-based snake. In the cavity case, please note that it gets stuck to a local minima.}
		\label{fig: snake_ref}
	\end{figure}
	
	Among all active contour models, snake\cite{Kass1988} is the most intuitive image segmentation model. It is based on the concept of a deformable curve or surface that can be iteratively adjusted to fit the edges or boundaries of an object in an image. The contour is driven by internal energy, which represents the smoothness of the curve, and external energy, which is derived from image features such as intensity or texture. By minimizing the total energy, the contour can accurately delineate the object boundaries, thus facilitating object recognition and tracking. For example, refer to Fig.\ref{fig: snake_ref}. At the start, the snake (or deformable contour) has maximum energy, and following the minimization process, it converges to the object's boundary. In mathematical terms, snakes detect object boundaries by minimizing edge-based energy functionals, achieved through the solution of the Euler-Lagrange equations. 
	
	Despite being very intuitive, snakes suffer from the following numerical issues \cite{sapiro2006geometric}:
	\begin{enumerate}
		\item They are sensitive to initialization.
		\item They don't work well with noisy data, which can be a significant issue in real-world scenarios where data may be incomplete or corrupted.
		\item Use of snake require sophisticated numerical methods \cite{Prince1997,Xie2008,Wang2009,COHEN1991211} to solve Euler-Lagrange equations.
		\item Hyperparameter tuning can be challenging, as selecting optimal parameters for the model can be time-consuming and require significant expertise.
		\item Snakes can't handle topology changes, which occur when objects in the image intersect or touch.
		\item It is difficult to incorporate prior domain knowledge in the form of energy functionals, which can limit the model's ability to leverage existing domain knowledge.
	\end{enumerate} 
	
	In this work, we propose PICS (Physics Informed Contour Selection) that integrates snakes and PINNs in a manner that capitalizes on the strengths of both approaches while mitigating their weaknesses. To accomplish this, we modified the PINN approach by introducing the following novelties:
	\begin{itemize}
		\item \textit{Custom design the PINN hypothesis (network architecture) to effectively capture object boundaries}. PINNs use a deep neural network (see eq.\ref{eqn: pinn_hyp}) with a large number of parameters to approximate the solution, whereas PICS employs cubic splines (see eq.\ref{eqn: spline_hyp}) that can efficiently approximate any closed contour with only a few control knots. The use of a simplified architecture leads to significant speed ups as compared to traditional PINNs\cite{dwivedi2020physics,dwivedi2021distributed,biharmonic2020,normal2021}.  
		\item \textit{Assign control knots as a design variable.} In most deep neural networks, the weights are typically initialized randomly and do not have any physical significance. However, PICS is formulated in such a way that trainable weights are represented by the control knots of cubic splines. It gives it a clear physical meaning to weights that simplifies scaling and normalization steps. Similarly, the loss gradient in PICS can be physically interpreted as the \textit{force} on the control knots pushing them towards the direction of gradient descent.
		\item \textit{Minimize region-based loss instead of gradient-based loss functionals.} PICS enhances the stability of snakes towards noisy data by optimizing region-based energy functionals \cite{mumford1989optimal,chan2001active} instead of relying solely on edge-based energy functionals \cite{Kass1988,Prince1997,Xie2008,COHEN1991211}. This approach has never been attempted in traditional numerical or deep learning frameworks because there is no explicit differentiable function for the derivative of the region-based loss with respect to snake control knots. We are the first to address this issue by implicitly calculating the loss derivatives through finite difference methods. 
		\item \textit{Incorporate prior shape information via regularization terms.} Like the parent PINN, PICS can easily accommodate any prior information about the shape of the object via shape-based regularization terms.
		\item \textit{Exploit transfer learning for 3D segmentation.} Given multiple 2D slices of a 3D object, PICS exploits transfer learning to reuse the optimized weights (spline control knots) from the previous slice as the initial condition for segmenting the current slice. This modality allows the snake to quickly converge to the optimal segmentation for each slice, reducing the number of iterations required for convergence and the overall computational cost.
	\end{itemize}
	
	To demonstrate the effectiveness of PICS, we take an example from the field of medical image segmentation. Medical image segmentation is a critical step in disease diagnosis, treatment planning, and medical research\cite{bai2020population,mei2020artificial,kickingereder2019automated,wang2019benchmark}. This process involves locating the regions of interest from medical images, such as magnetic resonance imaging (MRI) or computerized tomography (CT) scans, which can be used to identify and diagnose abnormalities, tumors, and other medical conditions. In recent years, medical image segmentation has been revolutionized by the rapid development of deep learning algorithms for computer vision\cite{badrinarayanan2017segnet,long2015fully, ronneberger2015u}. These algorithms have shown impressive results\cite{litjens2017survey,shen2017deep,hesamian2019deep,li2018h,dolz2018hyperdense,haberl2018cdeep3m} in accurately segmenting medical images, thus improving the accuracy and efficiency of medical diagnosis and treatment. However, their success heavily depends on the quality and quantity of the training data\cite{rahimi2021addressing,lecun2015deep,webb2018deep}. The problem is that acquiring high-quality medical images with labels (also called masks) requires expert interpretation \cite{https://doi.org/10.48550/arxiv.1711.08037}, and it is a labor-intensive and time-consuming process\cite{segars2013population,oktay2020evaluation}. Furthermore, the automatic tools are not trusted\cite{duran2021afraid} within the medical community. Therefore, PICS can fill a crucial research gap by serving as an intuitive tool for medical practitioners to generate rapid annotations, addressing the limitations associated with acquiring labeled medical images and improving the efficiency of the segmentation process. 
	
	The organization of this paper is as follows. We start with a brief review of snakes and PINN. Next, we list the objectives of the paper. In the Methods section, we describe the mathematical formulation of PICS. Then, we discuss the results in Results section. Finally, the conclusions of the paper are given at the end.
	
	\section{Brief Review of Snake and PINN} \label{Brief Review}
	\subsection{Brief Review of Snake Model}
	Consider Fig.\ref{fig: snake_ref} in which the snake (or deformable contour) has maximum energy at the start, and following the minimization process, it converges to the object's boundary. Mathematically, if the total energy of the snake is given by 
	\begin{equation}
		J=\frac{\alpha}{2}\intop_{0}^{1}\psi_{s}^{2}ds+\frac{\beta}{2}\intop_{0}^{1}\psi_{ss}^{2}ds+J_{ext}
	\end{equation} where $(s,t)$ denote space and time parameters respectively, $\psi$ denotes the parametric spline used for segmentation contour, $(\alpha, \beta)$ are the coefficients, the sum of first two terms denotes the internal energy ($J_{int}$) of the snake and $J_{ext}$ denotes external energy.
	
	Then, the motion of the snake is governed by the following PDE:
	\begin{equation}
		\frac{\partial\vv{\psi}}{\partial t}=\alpha\frac{\partial^{2}\vv{\psi}}{\partial s^{2}}+\beta\frac{\partial^{4}\vv{\psi}}{\partial s^{4}}-\nabla J_{ext}
	\end{equation}

	The internal energy controls the smoothness of snake and it is independent of the data. However, $J_{ext}$ is an image dependent, edge-based functional. For example, if $I$ is the image, then a simple gradient-based $J_{ext}$ could be $J_{ext}=-\int_{0}^{1}|\nabla I|^{2}ds$. For such functionals, a differentiable function for the gradient of $J_{ext}$ with respect to control knots cannot be found. However, if $J_{ext}$ is a region-based functional (for example, refer equation \ref{eq:cv}), then the expression for  $\nabla J_{ext}$ with respect to control knots cannot be directly found. If the object has weak gradients and the image is noisy, the denoising also removes the object boundary. In such cases, region-based loss functions are beneficial, but the traditional snake framework is not suitable for their implementation.
	
	\subsection{Brief Review of PINNs}
	\begin{figure}[ht!]
		\centering
		\includegraphics[width=0.65\linewidth]{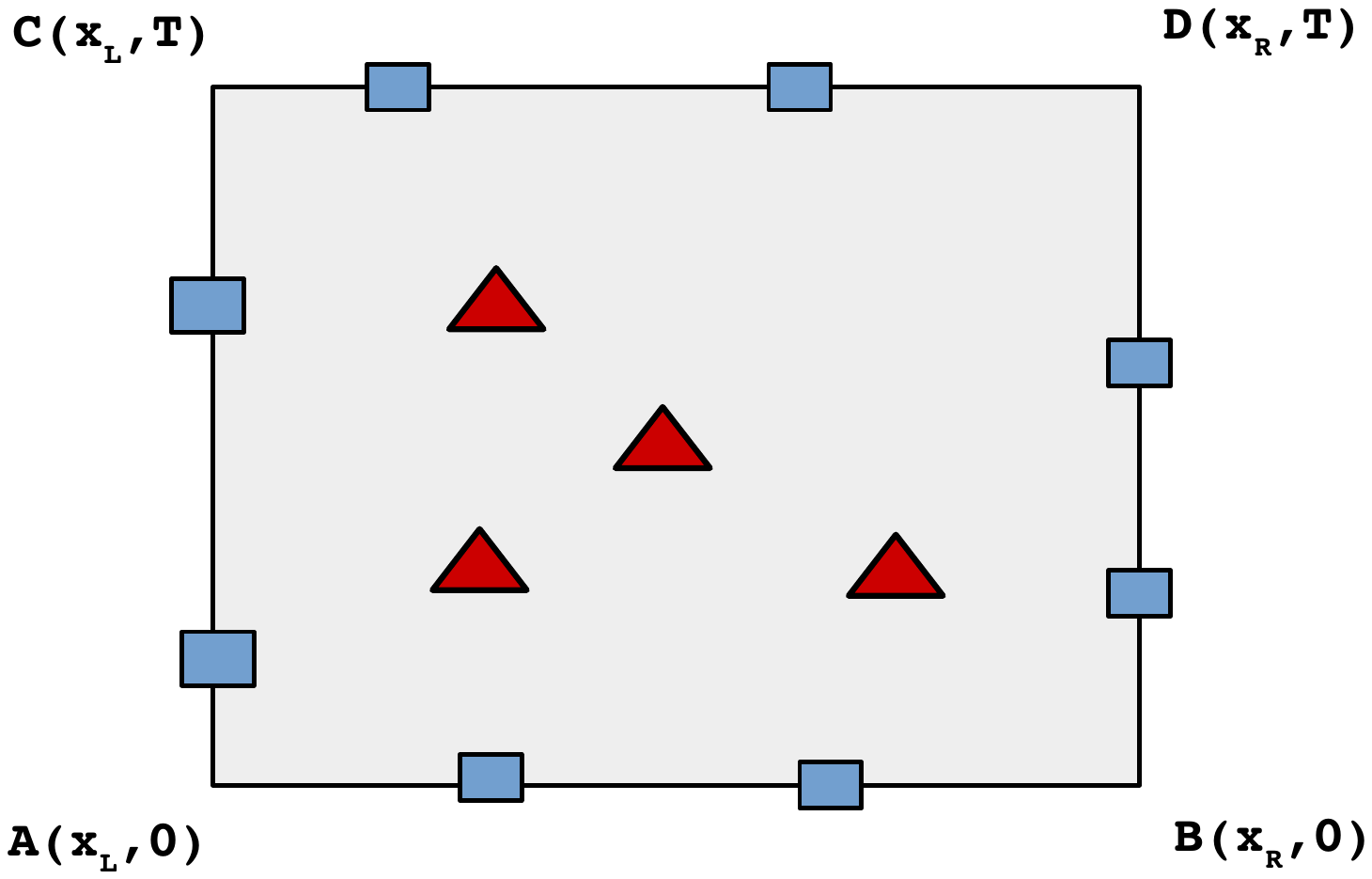}
		\caption{Distribution of collocation (red triangles) and boundary points (blue rectangles) in the computational domain. }
		\label{fig: distribution}
	\end{figure}
	In a typical PINN, the solution of PDE is approximated by a deep neural network. The training data, which consists of collocation and boundary points (see Fig.\ref{fig: distribution}), are randomly distributed in the computational domain. For example, consider the following one-dimensional (1D) unsteady PDE. 
	
	\begin{equation}
		\frac{\partial}{\partial t}u(x,t)+\mathcal{\mathcal{\mathscr{N}}}u(x,t)=R(x,t),(x,t)\in\Omega
		\label{eqn:PDE}
	\end{equation}
	\begin{equation}
		u(x,t)=B(x,t),(x,t)\in\partial\Omega,
		\label{eqn:BC}
	\end{equation}
	
	\begin{equation}
		u(x,0)=F(x),x\in(x_{L},x_{R}),
		\label{eqn:IC}
	\end{equation}
	where $\mathcal{\mathscr{N}}$ is a nonlinear differential operator and $\partial\Omega$ is the boundary of the computational domain $\Omega$. We approximate $u$ with a $n$-layered deep neural network $\psi$ such that 
	\begin{equation}
		\psi=\psi(z;W_{1},W_{2}...W_{n},b_{1},b_{2},...b_{n})=W_{n}(...(\phi(W_{2}(\phi(W_{1}z+b_{1}))+b_{2}))...)+b_{n}
		\label{eqn: pinn_hyp}
	\end{equation}
	where $z=[x,t]^{T}$denote sampling points, $(W_{i},b_{i})$ denote model parameters and $\phi$ denotes nonlinearity. For PINNs, $z$ are randomly selected, but after selection, they remain fixed. If we denote the errors in approximating the PDE, BCs, and IC by $\vv{\xi}_{f}$, $\vv{\xi}_{bc}$ and $\vv{\xi}_{ic}$ respectively. Then, the expressions for these errors are as follows:
	
	\begin{equation}
		\vv{\xi}_{f}=\frac{\partial\vv{\psi}}{\partial t}+\mathcal{\mathcal{\mathscr{N}}}\vv{\psi}-\vv{R},\;on\;(\vv{x}_{f},\vv{y}_{f})
		\label{eqn:err_PDE}
	\end{equation}
	\begin{equation}
		\vv{\xi}_{bc}=\vv{\psi}-\vv{B},\:(\vv{x}_{bc},\vv{t}_{bc})_{side\:faces}
		\label{eqn:err_BC}
	\end{equation}
	
	\begin{equation}
		\vv{\xi}_{ic}=\vv{\psi}(.,0)-\vv{F},\:(\vv{x}_{bc},\vv{t}_{bc})_{bottom\:face}
		\label{eqn:err_IC}
	\end{equation}
	For shallow networks, $\frac{\partial\vv{\psi}}{\partial t}$ and $\mathcal{\mathcal{\mathscr{N}}}\vv{\psi}$ can be determined using hand calculations. However, for deep networks, we have to use finite difference methods or automatic differentiation \cite{baydin2018automatic}. The latter is preferred for its computational efficiency. We can recast the PDE, BC, IC system to an optimization problem by minimizing an appropriate loss function. The loss function $J$ to be minimized for a PINN is given by
	\begin{equation}
		J=\frac{\vv{\xi}_{f}^{T}\vv{\xi}_{f}}{2N_{f}}+\frac{\vv{\xi}_{bc}^{T}\vv{\xi}_{bc}}{2N_{bc}}+\frac{\vv{\xi}_{ic}^{T}\vv{\xi}_{ic}}{2N_{ic}},
	\end{equation}
	where $N_{f}$, $N_{bc}$, and $N_{ic}$ refer to the number of collocation points, boundary condition points in left and right faces, and initial condition points at the bottom face, respectively. We can see that we have chosen a least square loss function. Now, any gradient based optimization routine may be used to minimize $J$.  
	
	\section{Objectives} \label{Objectives}
	The objective of the paper is to demonstrate the effectiveness of the PICS in performing image segmentation in both 2D and 3D settings, with and without prior knowledge of the object's shape. Specifically speaking,
	\begin{enumerate}
		\item Given a 2D image, perform segmentation with and without prior shape information.
		\item Given a 3D image, perform segmentation with and without prior shape information.
		\item Discuss hyperparameter tuning for simple and complex images.
	\end{enumerate}
	Furthermore, the paper aims to evaluate the performance of the method against labeled data to assess its accuracy and reliability.
	\section{Methods}\label{Methods}
	\begin{figure}[ht!]
		\centering
		\includegraphics[width=0.9\linewidth]{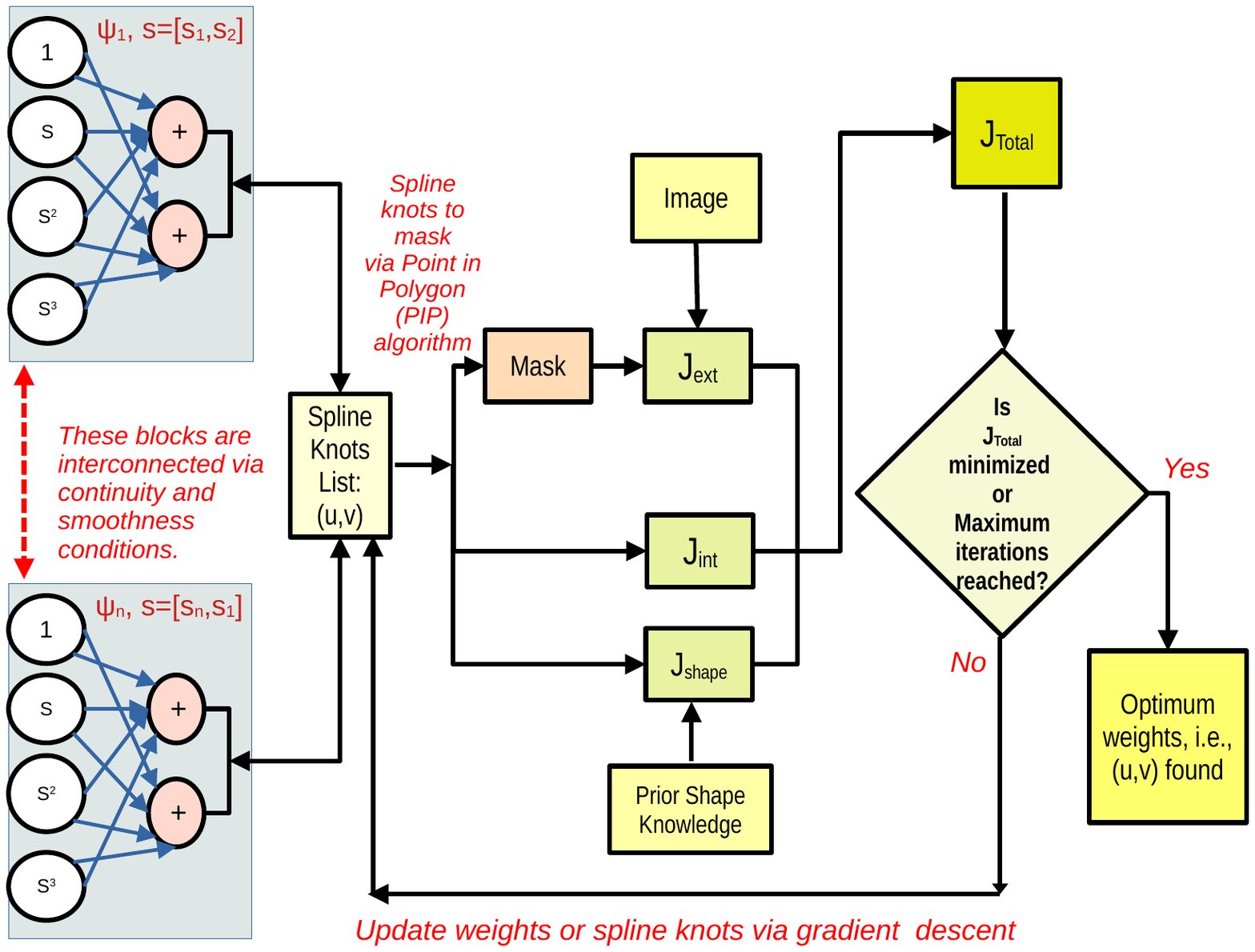}
		\caption{Overview of PICS algorithm.}
		\label{fig:PICS_Overview}
	\end{figure}
	Figure \ref{fig:PICS_Overview} shows the overall flowchart of PICS. In this section, we will describe its individual components, i.e., (a) PICS hypothesis, (b) the Chan-Vese loss function, (c) optimization,(d) the prior shape-based loss term, and (e) the operation performance index (OPI)--a metric to monitor the optimization performance of PICS.  
	\subsection{PICS Hypothesis}
	We approximate the solution in PICS, i.e., object boundary with a parametric spline. The expression of parametric spline $\vv{\psi}$ is given by
	
	\begin{equation}
		\vv{\psi}(s)=\begin{cases}
			\begin{array}{c}
				\vv{\psi}_{1}\\
				\vv{\psi}_{2}\\
				...\\
				\vv{\psi}_{n}
			\end{array} & \begin{array}{c}
				s_{1}<s<s_{2}\\
				s_{2}<s<s_{3}\\
				...\\
				s_{n}<s<s_{1}
		\end{array}\end{cases}
		\label{eqn: spline_hyp}
	\end{equation}
	
	where the local cubic spline $\vv{\psi}_{i}(s_{i}<s<s_{i+1})$ is given by 
	
	\begin{equation}
		\vv{\psi_{i}}(s)=\left\{ \begin{array}{c}
			u_{i}(s)\\
			v_{i}(s)
		\end{array}\right\} =\left\{ \begin{array}{c}
			a_{i}s^{3}+b_{i}s^{2}+c_{i}s+d_{i}\\
			e_{i}s^{3}+f_{i}s^{2}+g_{i}s+h_{i}
		\end{array}\right\}
	\end{equation}
	In the given equation, $s$ is a parameter that varies from 0 to 1. $[u_{i},v_{i}]'$ denotes the spatial coordinates of the local spline $\psi_{i}$. For given $[u_{i},v_{i}]'$ , the coefficients of local splines, denoted by $[a_{i},b_{i},...,h_{i}]'$, are computed by satisfying the conditions of continuity, smoothness, and periodicity. 
	
	It is important to note that unlike in PINN (recall eq. \ref{eqn: pinn_hyp}), these spline coefficients are not directly considered trainable weights in PICS. \textit{In PICS, the weights are the sampling points or the control knots themselves, i.e.,} $[u_{i},v_{i}]'$. With respect to Fig. \ref{fig:PICS_Overview}, we do not start with the left most blocks which represent local cubic splines. Our starting point is the list of spline knots. The spline coefficients are back calculated by a simple matrix operation. This novelty brings physical interpretability to the trainable parameters which otherwise don't have any direct physical significance. 
	
	Furthermore, PICS does not require any batch normalization as the physical constraints of continuity, smoothness, and periodicity, i.e., scale its weights,  
	\begin{enumerate}
		\item Continuity: $\vv{\psi}_{k-1}(s_{k})=\vv{\psi}_{k}(s_{k})=\left\{ \begin{array}{c}
			\tilde{u}(s_{k})\\
			\tilde{v}(s_{k})
		\end{array}\right\} $
		\item Smoothness: $\frac{d}{ds}\vv{\psi}_{k-1}(s_{k})=\frac{d}{ds}\vv{\psi}_{k}(s_{k})$
		and $\frac{d^{2}}{ds^{2}}\vv{\psi}_{k-1}(s_{k})=\frac{d^{2}}{ds^{2}}\vv{\psi}_{k}(s_{k})$
		\item Periodicity: $\frac{d}{ds}\vv{\psi}_{n}(s_{n})=\frac{d}{ds}\vv{\psi}_{1}(s_{1})$
		and $\frac{d^{2}}{ds^{2}}\vv{\psi}_{n}(s_{n})=\frac{d^{2}}{ds^{2}}\vv{\psi}_{1}(s_{1})$
	\end{enumerate}  
	
	\subsection{Loss Function and Optimization}
	The formula for the region-based \cite{mumford1989optimal,chan2001active} loss function is given by
	\begin{equation}
		J=\alpha J_{\psi_{s}}+\beta J_{\psi_{ss}}+\mu J_{cv}
	\end{equation}
	where 
	\begin{equation}
		J_{\psi_{s}}=\sum_{i=1}^{i=N}\frac{1}{N}\left(\left(\frac{d\widetilde{u}}{ds}\right)^{2}+\left(\frac{d\widetilde{v}}{ds}\right)^{2}\right)_{i}
	\end{equation}

	\begin{equation}
		J_{\psi_{ss}}=\sum_{i=1}^{i=N}\frac{1}{N}\left(\left(\frac{d^{2}\widetilde{u}}{ds^{2}}\right)^{2}+\left(\frac{d^{2}\widetilde{v}}{ds^{2}}\right)^{2}\right)_{i}
	\end{equation}
	
	\begin{dmath}\label{eq:cv}
		J_{cv}=\sum_{q=1}^{q=N_{y}}\sum_{p=1}^{p=N_{x}}\left(\left(I(p,q)-\mu_{in}\right)\chi(p,q)\right)^{2}+\gamma\left(\nabla I(p,q)\chi(p,q)\right)^{2}+\left(\left(I(p,q)-\mu_{out}\right)(1-\chi(p,q))\right)^{2}
	\end{dmath}
	In the above expressions, $I$ denotes the image, $(\widetilde{u},\widetilde{v})$ denotes spline knots,  and $N$ denotes the number of spline control knots. $\chi$ denotes a characteristic function or mask that is generated by repeated geometric queries, that is, given a single polygon through spline knots and a sequence of query points (grid points), find if the query point lies inside or outside the polygon using point in polygon algorithms \cite{shimrat1962algorithm}. $\mu_{in},\mu_{out}$ denote the average pixel value of the image within and outside the spline contour. $N_{x},N_{y}$ denote number of pixels in $x$ and $y$ direction respectively. With respect to Fig. \ref{fig:PICS_Overview}, the external energy term of the loss is $J_{ext}=\mu J_{cv}$, and the internal energy term is $J_{int}=\alpha J_{\psi_{s}}+\beta J_{\psi_{ss}}$. The hyperparameter $\gamma$ aims to make the pixel intensities inside the contour more uniform. 
	
	While doing the weight update by gradient descent, the derivative of loss with respect to weights is compulsory. For example, The expression for weight update using gradient descent is given by 
	\begin{equation}
		w=w_{old}-\lambda\frac{\partial J}{\partial w_{old}}
	\end{equation}
	where $w$ denotes weight or spline control knots, $\lambda$ is the learning rate and $\frac{\partial J}{\partial w}$ is the loss gradient. The loss gradient cannot be calculated directly or even by automatic differentiation because there is no explicit differentiable function that maps control knots with mask. Therefore to calculate derivatives, we use central difference scheme. This is the advantage of minimizing the energy functional instead of using PINN-like PDE residual, as it relaxes the differentiability requirements. For faster convergence and adaptive learning rate, we have used Adam \cite{https://doi.org/10.48550/arxiv.1412.6980} optimizer for our numerical experiments.  
	
	\subsection{Prior Shape-Based Loss Term}
\begin{figure}[ht!]
		\centering
		\includegraphics[width=0.9\linewidth]{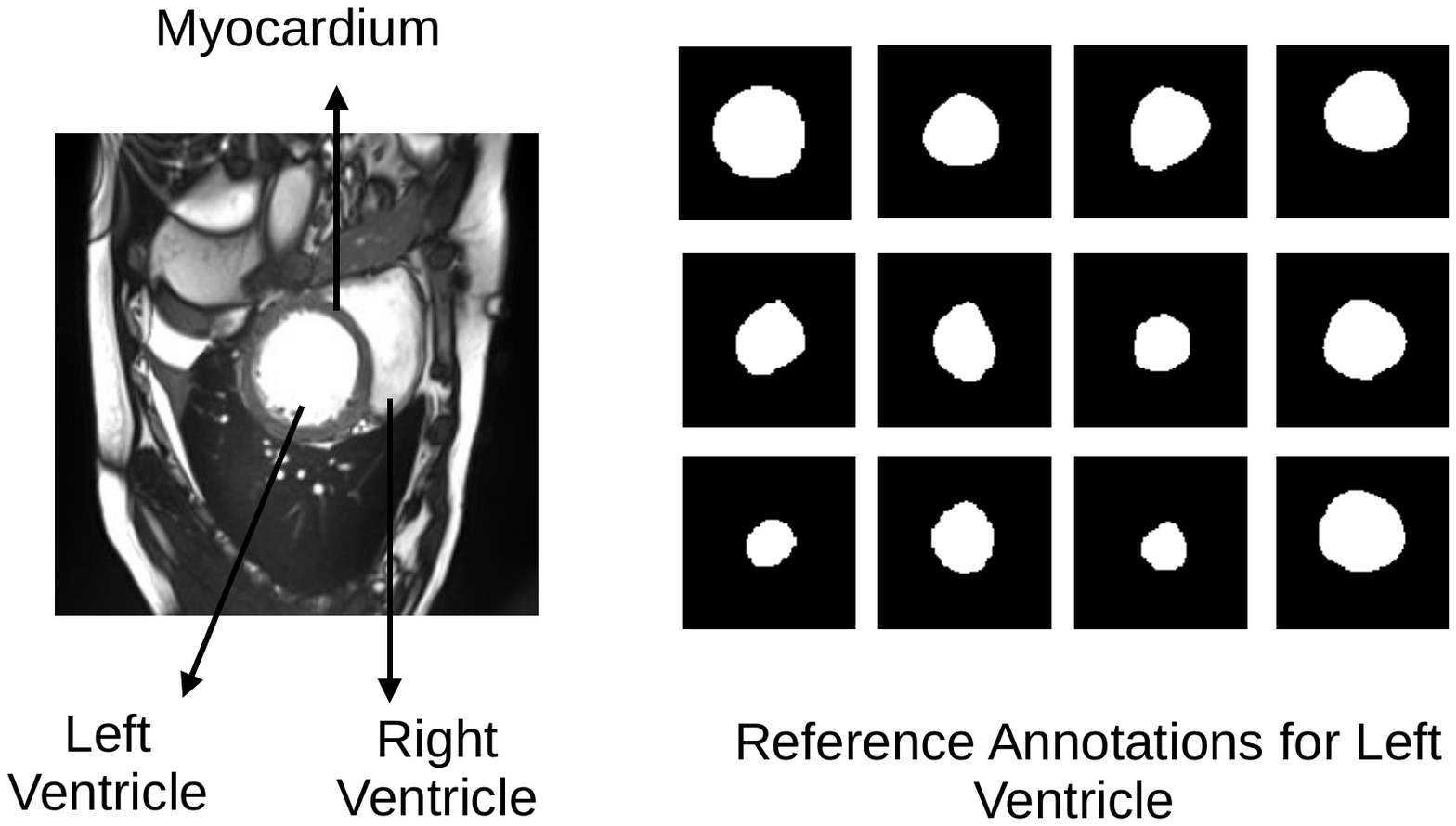}
		\caption{Description of the cardiac dataset and sample annotations for the left ventricle.}
		\label{fig: Shape_Prior}
	\end{figure}

	This paper will use PICS to generate annotations for the left ventricle in the cardiac MRI scan images. A representative cardiac MRI scan shown in the left-hand side of Figure \ref{fig: Shape_Prior} is composed of three main parts: left ventricle, right ventricle, and myocardium. In clinical applications of cardiac left ventricle (LV) segmentation, it is desirable to include the cavity, trabeculae, and papillary muscles, which collectively form a convex shape, as shown by the right-hand side of Figure \ref{fig: Shape_Prior} where some reference annotations for left ventricle are depicted. However, trabeculae and papillary muscles have similar intensities to the myocardium, which can cause segmentation algorithms to incorrectly classify them as myocardium.
	
	The problem here is to find a way to accommodate medical domain knowledge with a purely data or image-driven algorithm. To address this challenge, Shi and Li \cite{SHI2021109} developed a method that preserves the convexity of the left ventricle by controlling the curvature in the level set framework. Similarly, in PICS, we propose a new loss term that preserves convexity in the snake framework. This loss term is expressed as follows:
	\begin{equation}
		J_{shape}=\sigma\sum_{i=1}^{i=N}\frac{1}{N}\kappa^{2}=\sigma\sum_{i=1}^{i=N}\frac{1}{N}\left(\frac{\tilde{v}_{ss}\tilde{u}_{s}-\tilde{u}_{ss}\tilde{v}_{s}}{(\tilde{u}_{s}^{2}+\tilde{v}_{s}^{2})^{\frac{3}{2}}}\right)^{2}
	\end{equation}where $\kappa$ denotes the curvature of contour, $N$ denotes number of spline control knots and $\sigma$ is a hyperparameter. This penalty term ensures that the shape of the predicted boundary remains convex-shaped. Please note that such an information about the shape of the object is not always available. In those cases, as Fig. \ref{fig:PICS_Overview} shows, PICS works with just $J_{int}$ and $J_{ext}$. 
	
	\subsection{Operation Performance Index (OPI)}
	The region-based loss function is comprised of both shape regularization and external energy (or mean square error) terms. During optimization, our aim is to reduce the total loss. However, if the relative strengths of the shape regularization and MSE terms need to be appropriately balanced, the solution may get stuck in a local minimum. 
	
	To prevent this issue, we introduce a new performance evaluation metric known as the Operation Performance Index (OPI). A value of one for OPI indicates that PICS is moving in the right direction, while a value below a predetermined threshold (e.g., 0.8) indicates that parameter adjustments are necessary. However, what constitutes the "right direction"? In the best-case scenario, both external and internal energies should drop as optimization proceeds. However, if that is not possible, the external energy should always drop whether shape regularization loss drops or not. This idea is mathematically contained in the following formula for OPI:
	
	\begin{equation}
		OPI_{k}=1-\frac{<\vv{\theta_{k}},\vv{P_{k}}>}{2\sum\vv{\theta_{k}}}
	\end{equation}
	where
	\[
	\vv{P_{k}}=sign(\triangle J_{int}(k-w+1:k))-sign(\triangle J_{ext}(k-w+1:k))
	\]
	\[
	\vv{\theta_{k}}=\frac{exp([1,1+d,...1+id,...2])}{\sum_{i=0}^{i=w-1}exp(1+id)},d=\frac{1}{w-1}
	\]
	$k$ , $w$ and $<>$ represent iteration number, iteration window size, and dot product, respectively. $\triangle J$ represents vector of difference in $J$. The exponential smoothing term ensures that the recent values are given more weightage. 
	
	OPI can also be used for hyperparameter tuning. For instance, the values of the hyperparameters $\alpha$ and $\beta$ can be determined through trial and error. But, the third parameter, $\mu$, which is initially set to 1e3, can be adjusted using OPI. When OPI falls below the threshold, the update rule for $\mu$ is given by
	\begin{equation}
		\mu=\mu+2^{log_{10}\left(J_{ext}/J_{int}\right)}
	\end{equation}
	However, we do not continue adjusting $\mu$ indefinitely. We stop adjusting $\mu$ once the order of $J_{ext}$ becomes more than 1e4 times that of $J_{int}$ to prevent the snake from becoming too loose. 
	
	When examining the total loss history alone, it is difficult to determine whether the optimization is progressing correctly. However, looking at the OPI trend, we can check if its value is very low or wildly fluctuating between 0 and 1. Based on this information, the hyperparameters may be adjusted. We will cover the application of this idea in the next section.
	
	We summarize this section with two tables. Table \ref{tab:PINN_Vs_PICS} lists the similarities and differences between PICS and PINNs, and Table \ref{tab:Snakes_Vs_PICS} lists the problems with snake model and their remedies in PICS. 
	\begin{table}[ht!]
		\begin{centering}
			\begin{tabular}{|p{3cm}|p{3.5cm}|p{3.5cm}|}
				\hline 
				\textbf{Property} & \textbf{PINN} & \textbf{PICS} \tabularnewline
				\hline 
				\hline 
				Basis function & Deep neural network (nonlinear) & Cubic splines (linear)\tabularnewline
				\hline 
				Parameters & Weights and biases (no physical interpretation) & Control knots (interpretable) \tabularnewline
				\hline
				The gradient of loss with respect to parameters & No physical interpretation & Force on the control knots \tabularnewline
				\hline
				Training Points & Scattered points (mesh-free) & Scattered points (mesh-free) \tabularnewline
				\hline
				PDE embedding & PDE as loss function & PDE functional as loss function \tabularnewline
				\hline
				Optimization & Gradient-based & Gradient-based \tabularnewline
				\hline
			\end{tabular}
			\par\end{centering}
		\caption{\label{tab:PINN_Vs_PICS} Comparison between PINN and PICS.}
	\end{table}
	\begin{table}
		\centering
		\begin{tabular}{|l|p{5cm}|p{5cm}|}
			\hline
			\textbf{S.No.} & \textbf{Snake problem} & \textbf{PICS solution} \tabularnewline
			\hline
			1 & Sensitive to initialization & Physically consistent initialization by humans \tabularnewline
			\hline
			2 & Sensitive to noisy data & Robust to noisy data due to use of region-based functionals\tabularnewline
			\hline
			3 & Requires sophisticated algorithms to solve Euler-Lagrange equations & Requires simple gradient-descent for all kind of loss functionals\tabularnewline 
			\hline
			4 & Difficult to incorporate shape-priors & Easy to incorporate shape-priors as loss functions  \tabularnewline
			\hline
			5 & Only forward problems & Forward as well as inverse problems \\
			\hline
		\end{tabular}
		\caption{\label{tab:Snakes_Vs_PICS} PICS solution for snake problems.}
	\end{table}

	\section{Results and Discussion} \label{Results and Discussion}
	In this section, we demonstrate the effectiveness of PICS in 2D and 3D segmentation with and without prior shape information. In all the cases, Adam optimizer \cite{https://doi.org/10.48550/arxiv.1412.6980} is used. All the experiments are conducted in Matlab R2022b environment running in a 12th Gen Intel(R) Core(TM) i7-12700H, 2.30 GHz CPU and 16GB RAM Asus laptop. For testing PICS, we have considered the following cases:
	\begin{enumerate}
		\item CT scan of enlarged ventricles of hydrocephalus patient (Case courtesy of Paul Simkin, Radiopaedia.org, rID: 30453). The source of data is: \url{https://radiopaedia.org/cases/obstructive-hydrocephalus}
		\item Synthetic image of a disk with distorted boundaries. This test case was also used by Xie et al. \cite{Xie2008}.
		\item Synthetic image of a cavity \cite{Prince1997,Xie2008,Wang2009}. This is a standard test case where traditional snake models have been observed to be unsuccessful in navigating through concavities.
		\item MRI scans of cardiac data of 100 patients from the ACDC dataset\cite{bernard2018deep} in the ED, i.e., End-Diastolic phase. Source:\url{https://www.creatis.insa-lyon.fr/Challenge/acdc/index.html}
	\end{enumerate}
	
	The first three tests can be assessed visually without the need for prior shape information. However, the last test, which uses the ACDC dataset, requires expert interpretation and therefore is evaluated by comparing the results with annotations provided by medical professionals. The intersection over union (IoU) is used as the evaluation metric to compare the results. IoU measures the similarity between two sets of data by dividing the size of their intersection by the size of their union. The formula for IoU is: $$IoU(A, B) = \frac{|A \cap B|}{|A \cup B|}$$where A and B are the two sets being compared. The IoU value ranges from 0 to 1, where 0 indicates no common elements between the sets and 1 indicates identical sets.
	
	\subsection{2D image segmentation without any prior shape information}
\begin{figure}[ht!]
		\centering
		\includegraphics[width=0.9\linewidth]{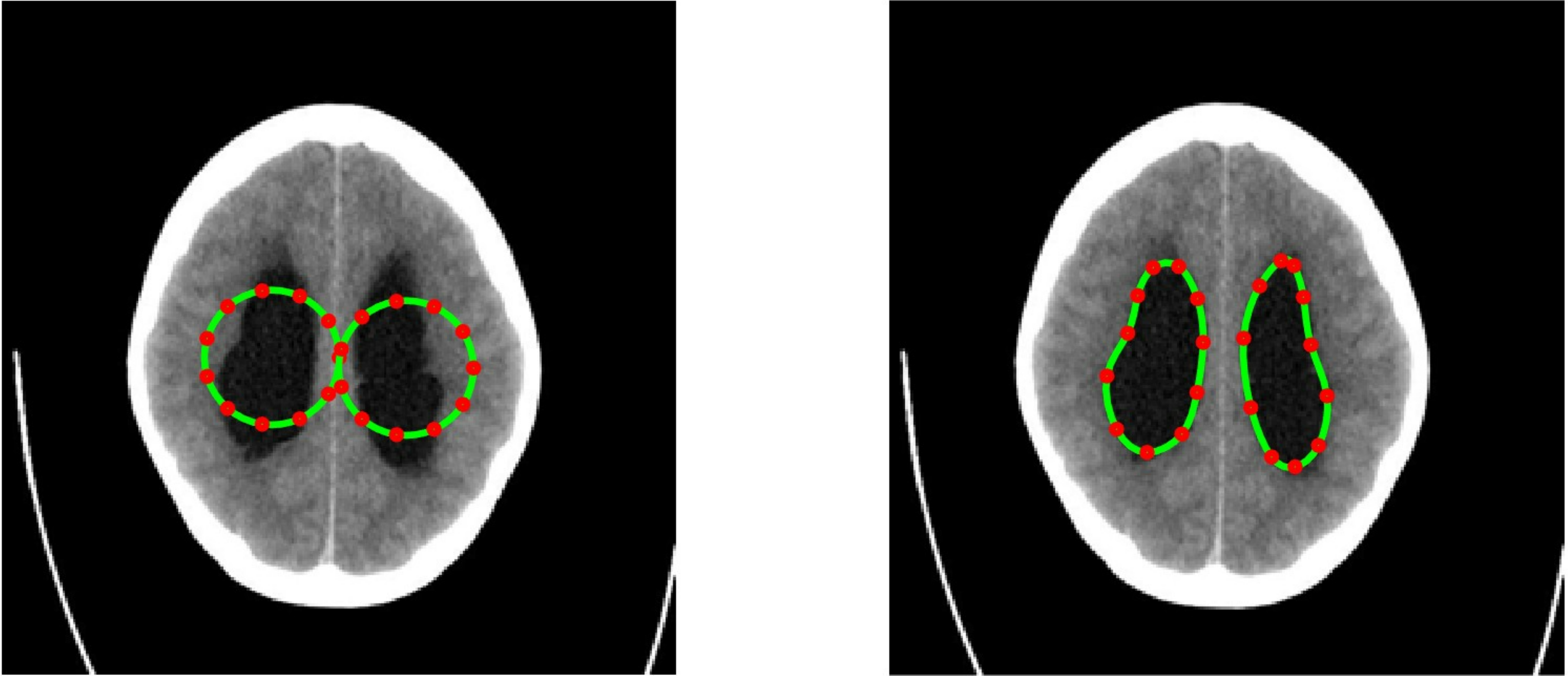}
		\caption{segmentation of enlarged ventricles of a hydrocephalus patient. Left: Initial weights, Right: Optimized weights.}
		\label{fig: 2D_Seg_HCF}
	\end{figure}

	In this case, we consider a CT scan of the enlarged ventricles of a hydrocephalus patient. Figure \ref{fig: 2D_Seg_HCF} shows the segmentation results. The CT scan shows two enlarged ventricles. The left-hand side of Figure \ref{fig: 2D_Seg_HCF} shows the contours initialized by the PICS user, and the right-hand side figure shows the optimized weights. By visual inspection, we can conclude that the results are satisfactory. The number of trainable parameters (i.e., two times the control knots) $N$ equals 22. The values of hyperparameters used are: ($\alpha$,$\beta$,$\mu$,$\gamma$,$\sigma$)=(5e-1,5e-2,1e3,0,0). The values of $\alpha$ and $\beta$ are fixed by trial and error. The value of $\mu$ is adaptive depending on the OPI discussed earlier. The time taken during the optimization process is close to a minute. Therefore, PICS shows good performance when dealing with images that contain a single target object or when the number of target objects is known. 
	
	\subsection{2D image segmentation with prior shape information}
\begin{figure}[ht!]
		\centering
		\includegraphics[width=0.9\linewidth]{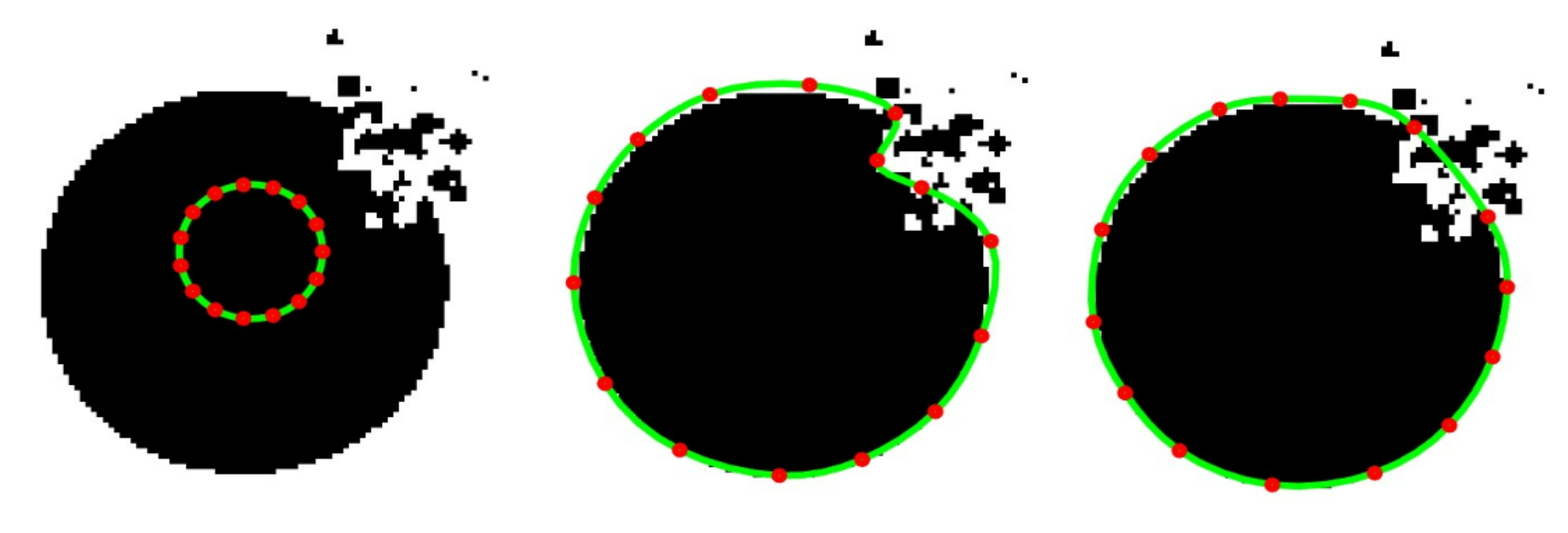}
		\caption{Effect of inclusion of convexity preserving loss term in 2D segmentation. Left: Initial weights, Middle: Optimized weights without shape prior, Right: Optimized weights with shape prior.}
		\label{fig: 2D_Seg_Shape_Prior}
	\end{figure}

	In this case, we consider the synthetic image of a disk with distorted boundaries. Figure \ref{fig: 2D_Seg_Shape_Prior} shows the segmentation results without and with shape prior. The synthetic disk has a partially distorted boundary. The left-hand side of Figure \ref{fig: 2D_Seg_Shape_Prior} shows the contours initialized by the PICS user. The middle figure shows the segmentation result without the shape loss term. The right-hand side figure shows the segmentation result with the shape loss term. The number of trainable parameters (i.e., two times the control knots) $N$ equals 30. The values of hyperparameters without shape loss term are: ($\alpha$,$\beta$,$\mu$,$\gamma$,$\sigma$)=(5e-1,1e-2,1e4,0,0) and with shape prior are ($\alpha$,$\beta$,$\mu$,$\gamma$,$\sigma$)=(5e-1,1e-2,1e4,0,1e8). The values of $\sigma$ are fixed by trial and error. Therefore, the inclusion of a shape loss term enables PICS to nearly recover the original shape of an object even when the boundary is distorted. 
	\subsection{3D image segmentation with or without prior shape information}
	\begin{itemize}
		\item \textit{Effect of inclusion of convexity preserving loss term in LV segmentation.} Figures \ref{fig: 2D_LV_Without_With_Shape_Prior_Knots} and \ref{fig: 2D_LV_Without_With_Shape_Prior_masks} demonstrate the impact of integrating a convexity-preserving loss term in the segmentation of the left ventricle for the ED (End-diastolic) case. As depicted in the middle figure, a purely data-driven segmentation algorithm is likely to fail in cases where trabeculae and papillary muscles have comparable intensities to the myocardium. However, by incorporating prior shape information that preserves convexity, PICS is able to accurately segment the left ventricle even in the presence of confusing muscles. The inclusion of the shape loss term results in an increase of the IoU score from 0.49 without the shape loss term to 0.87 with it. The values of hyperparameters without and with shape loss term are ($\alpha$,$\beta$,$\mu$,$\gamma$,$\sigma$)=(5e-1,1e-3,1e4,0,0) and (5e-1,1e-3,1e4,0,1e8) respectively. The number of trainable weights in both cases is 20. 
		
		\item \textit{Rapid 3D segmentation via transfer learning.} If $F$ represents a nonlinear transformation that takes image $I(x,y)$ and initial weights $\vv{w}$ as input and gives weight update as output. We can model the 3D segmentation process as follows:
		\begin{equation}
			\vv{w}^{[n]}=\vv{w}_{opt}^{[n-1]}+F(I(x,y)^{[n]},\vv{w}_{opt}^{[n-1]}),n=1,2,3,..
		\end{equation}
		where the weights are initialized by a single mouse click of the user, i.e.,
		\begin{equation}
			\vv{w}^{[0]}=\vv{w}_{user}
		\end{equation}
		It is mathematically equivalent to solving an unsteady PDE given by 
		\begin{equation}
			\frac{d\vv{w}}{dt}=F(I(x,y),\vv{w})
		\end{equation} 
		
		Figure \ref{Fig: 3D_Seg_LV} demonstrates the 3D segmentation process of the left ventricle in the PICS framework. The initialization process begins with a single click within the left ventricle on the first MRI image. Subsequently, the optimal PIC weights obtained from the previous image are transferred as the initial condition for the next image. The transfer learning accelerates the convergence of the PIC to its optimal value. This iterative process is continued for all the remaining images. Figure \ref{fig: Speed_Up} illustrates an example of the speed-up in convergence due to this transfer learning process.
		
		\item \textit{Performance evaluation on ACDC dataset.} Figure \ref{fig: 3D_Seg_LV_Final} shows the performance of PICS on all the hundred patients' data, with an average IoU of 0.88. The probability mass function is also shown. The number of trainable parameters for all the cases is the same, which is 20 parameters. However, the choice of hyperparameters is not the same for all cases. Based on the images in the ACDC dataset, they can be broadly categorized into three classes (see Figure \ref{fig: Three_Classes}): (1) Normal Case, (2) Indistinct Muscles, and (3) Very Thin Myocardium. PICS uses different hyperparameter settings for different image categories in the ACDC dataset. For the normal case, the hyperparameters have values of ($\alpha$,$\beta$,$\mu$,$\gamma$,$\sigma$)=(1e-1,1e-2,1e4,1e-5,1e7). For the indistinct muscles category, the value of $\sigma$ is increased by a factor of 10 while keeping all other hyperparameters fixed. Similarly, for the last class with very thin myocardium, the value of $\gamma$ is increased by a factor of 100-200 while keeping all other hyperparameters fixed. These hyperparameter settings are based on the observations from the ACDC dataset and have been found to provide good performance in their respective image categories. 
		
		\textit{Comparison with winners of ACDC challenge}. First and foremost, it's important to recognize that comparing PICS with a deep learning-based segmentation algorithm isn't entirely fair because PICS creates annotations (ground truth), while neural networks make predictions. The best IoU score(\url{https://www.creatis.insa-lyon.fr/Challenge/acdc/results.html})is 0.96 while ours is 0.88. Despite the fact that our result (IoU=0.88) places us at the bottom of the leaderboard, it's still a promising result when compared to traditional deep neural networks which are complex, with millions of parameters, requiring extensive training sessions lasting several hours and a substantial amount of high-quality data for effective training.
		
		The annotations generated by PICS can serve as a preliminary mask for the labeler, who can then refine them by correcting the positions of control knots as needed. From an algorithmic perspective, this refinement can be accomplished by fixing the satisfactory control knots or weights  as "non-trainable" and allowing the remaining ones to adjust. Within the scope of this discussion, the results obtained in our current study, which serves as an initial proof of concept, are very promising.
		\begin{figure}[ht!]
			\centering
			\includegraphics[width=0.9\linewidth]{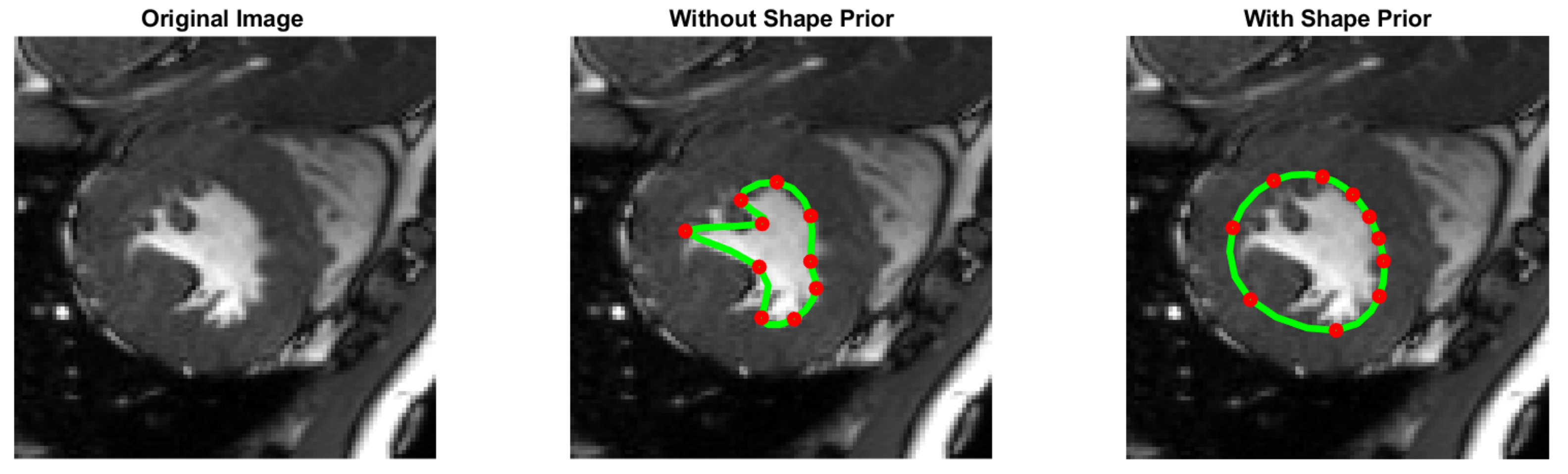}
			\caption{Effect of inclusion of convexity preserving shape prior in segmentation (control knots) of the left ventricle.}
			\label{fig: 2D_LV_Without_With_Shape_Prior_Knots}
		\end{figure}
		\begin{figure}[ht!]
			\centering
			\includegraphics[width=0.9\linewidth]{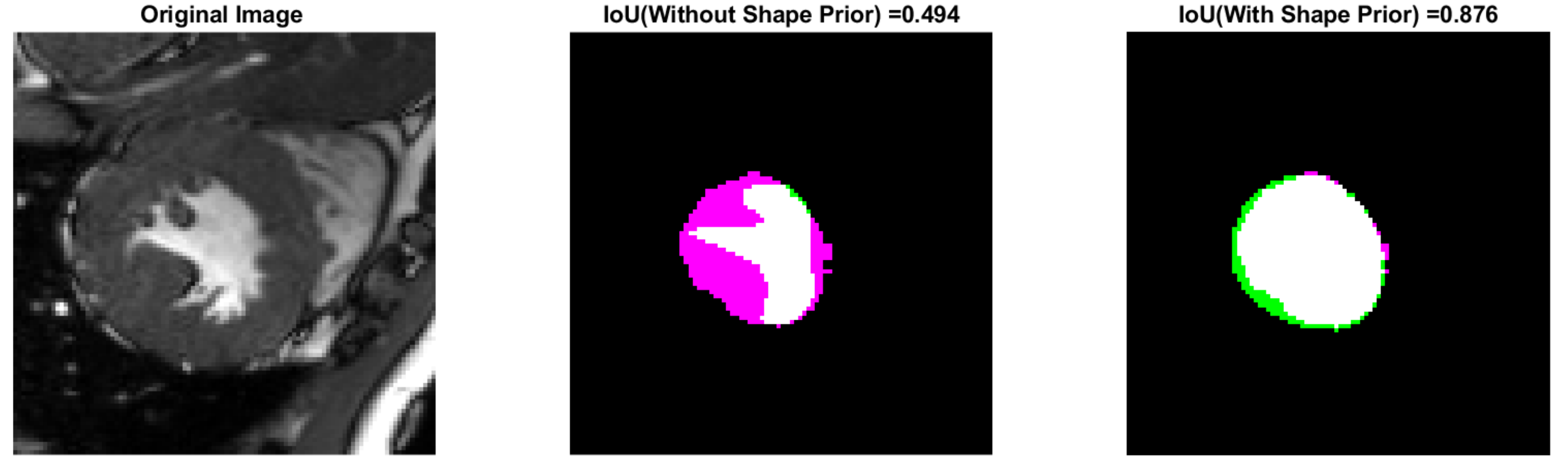}
			\caption{Effect of inclusion of convexity preserving shape prior in segmentation (masks) of the left ventricle.}
			\label{fig: 2D_LV_Without_With_Shape_Prior_masks}
		\end{figure}
		\begin{figure}[ht!]
			\centering
			\begin{minipage}[b]{\textwidth}
				\centering
				\begin{subfigure}[b]{0.45\textwidth}
					\centering
					\includegraphics[width=\linewidth]{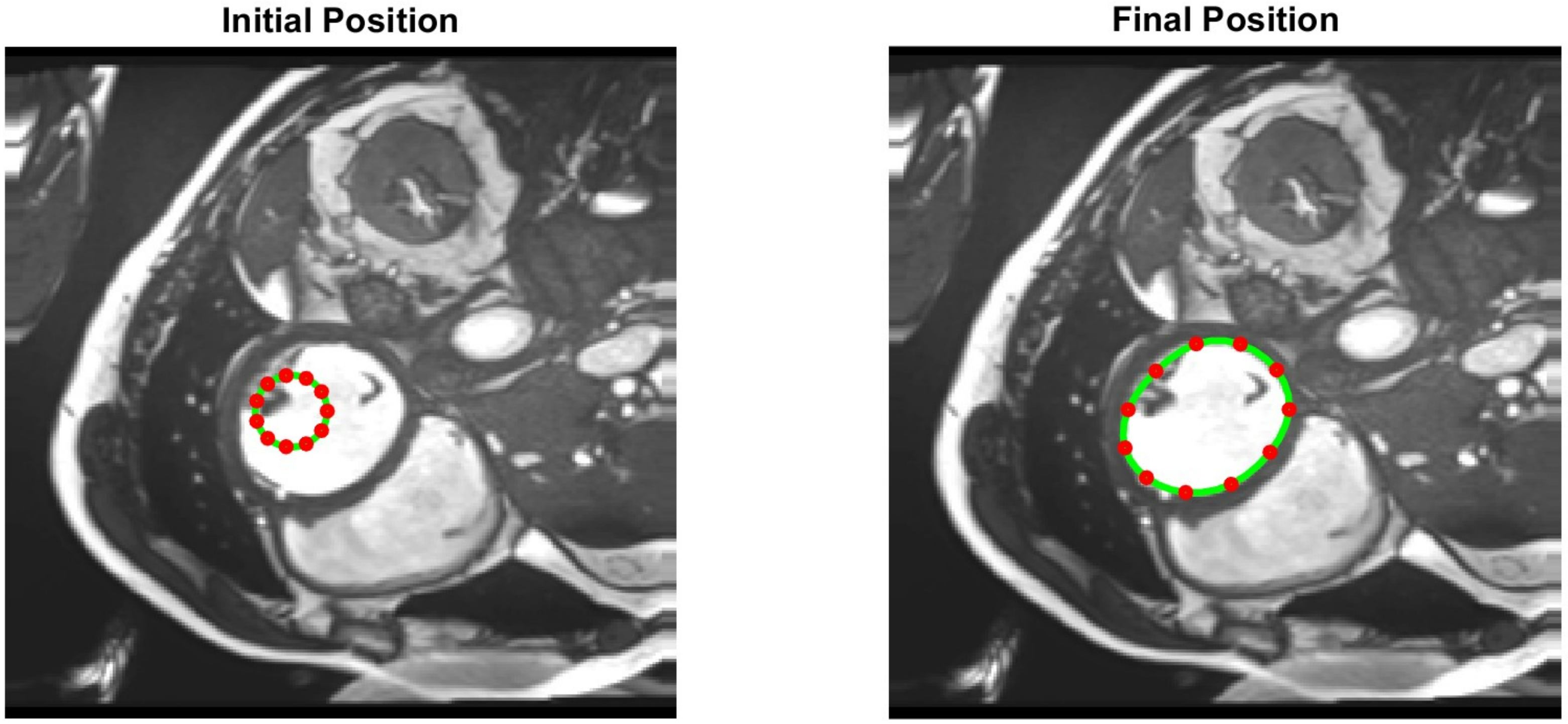}
					\caption{Image-1: User is asked to click within left ventricle for weights initialization. }
				\end{subfigure}
				\hspace{0.05\textwidth}
				\begin{subfigure}[b]{0.45\textwidth}
					\centering
					\includegraphics[width=\linewidth]{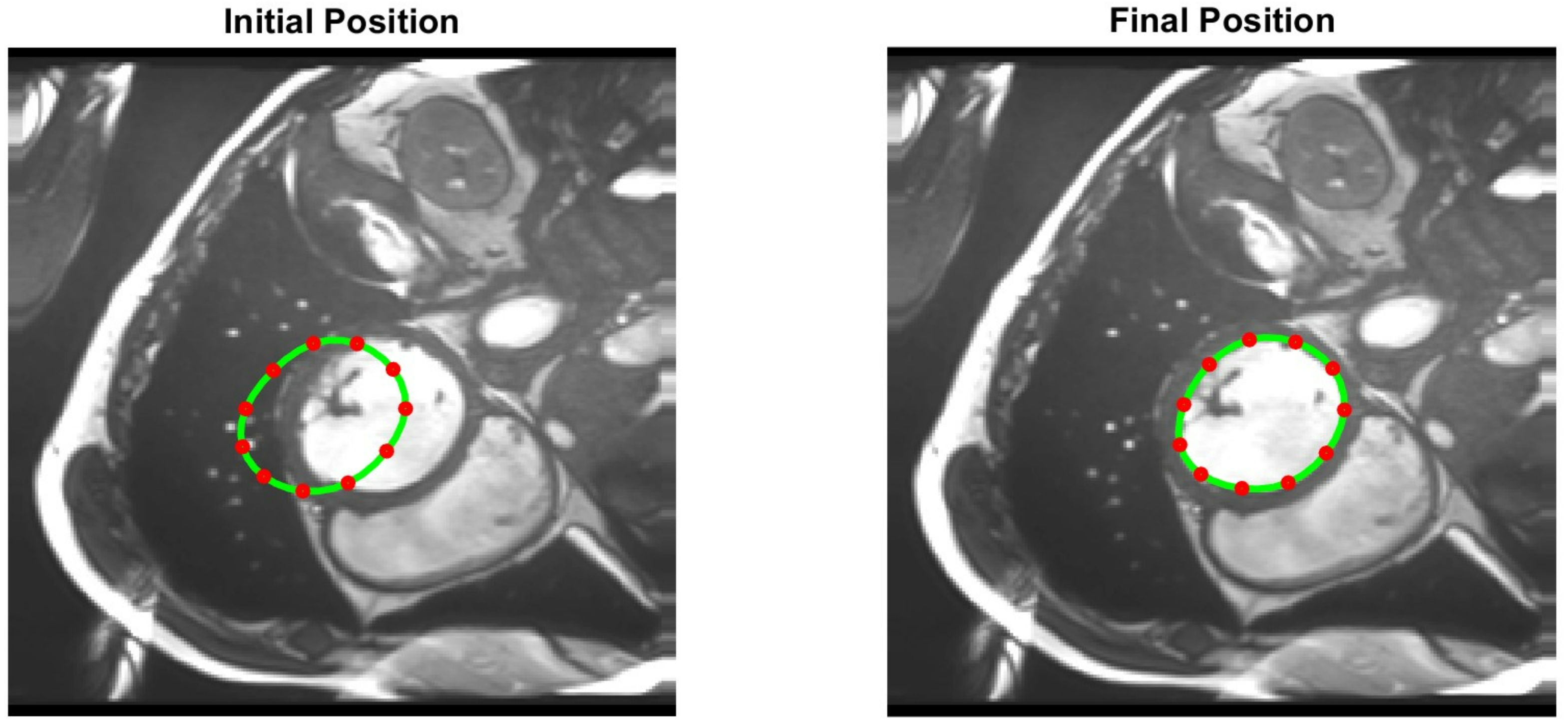}
					\caption{Image-2: Optimum weights of the first slice serve as initial weights for the second.}
				\end{subfigure}
				
				\vspace{0.05\textwidth}
				
				\begin{subfigure}[b]{0.45\textwidth}
					\centering
					\includegraphics[width=\linewidth]{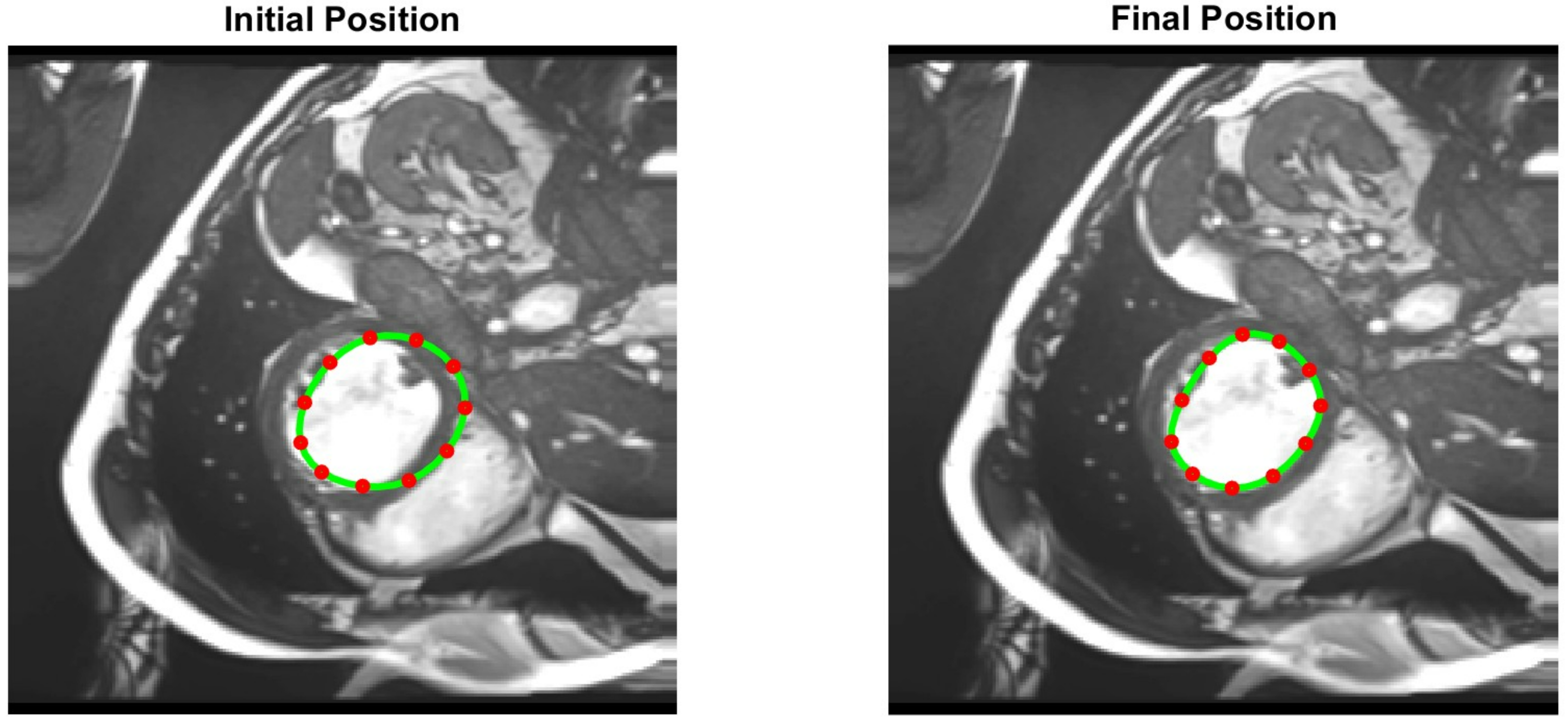}
					\caption{Image-3: Optimum weights of the second slice serve as initial weights for the third.}
				\end{subfigure}
				\hspace{0.05\textwidth}
				\begin{subfigure}[b]{0.45\textwidth}
					\centering
					\includegraphics[width=\linewidth]{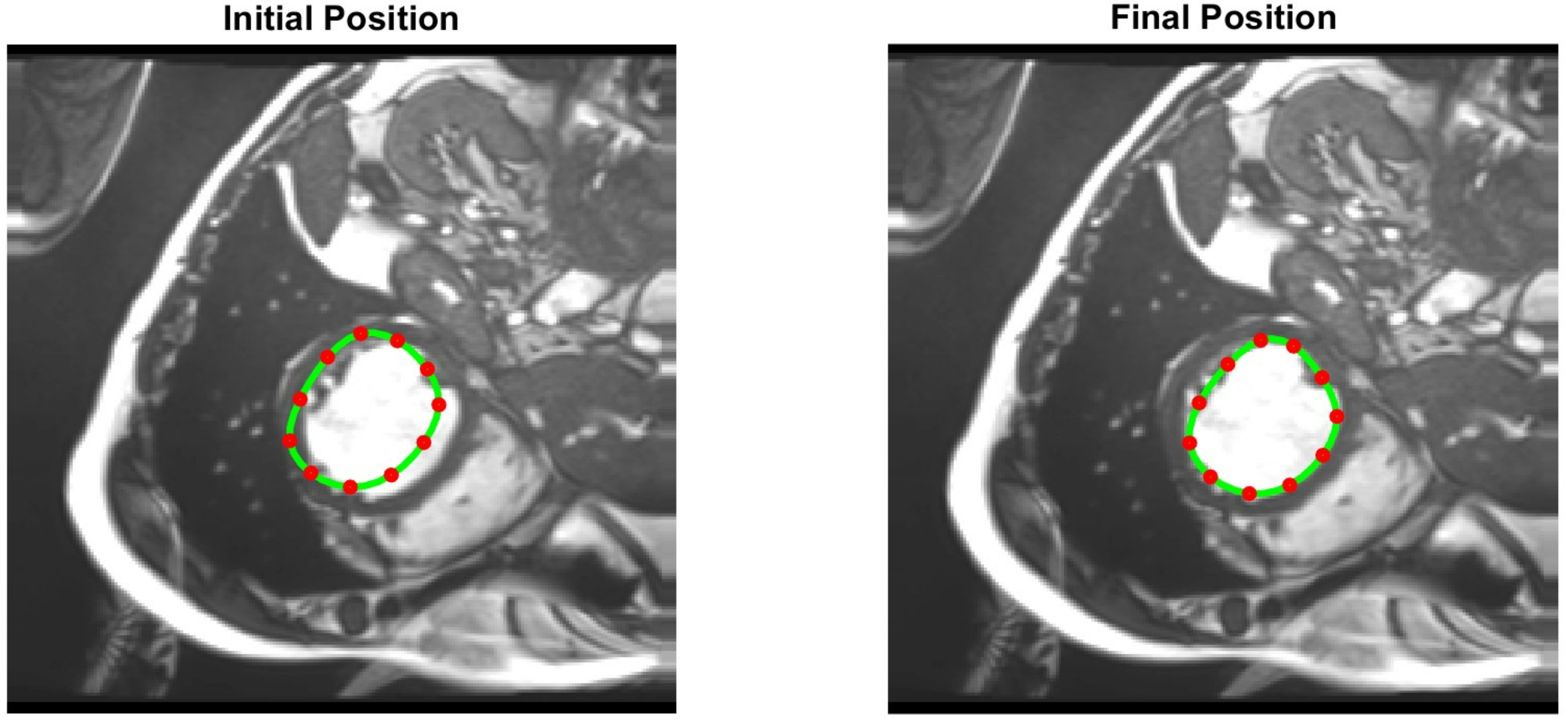}
					\caption{Image-4: Optimum weights of the third slice serve as initial weights for the fourth.}
				\end{subfigure}
			\end{minipage}
			\caption{\label{Fig: 3D_Seg_LV}3D segmentation of left ventricle with PICS using transfer learning (for ED phase).}
		\end{figure}
		
		\begin{figure}[ht!]
			\centering
			\includegraphics[width=0.6\linewidth]{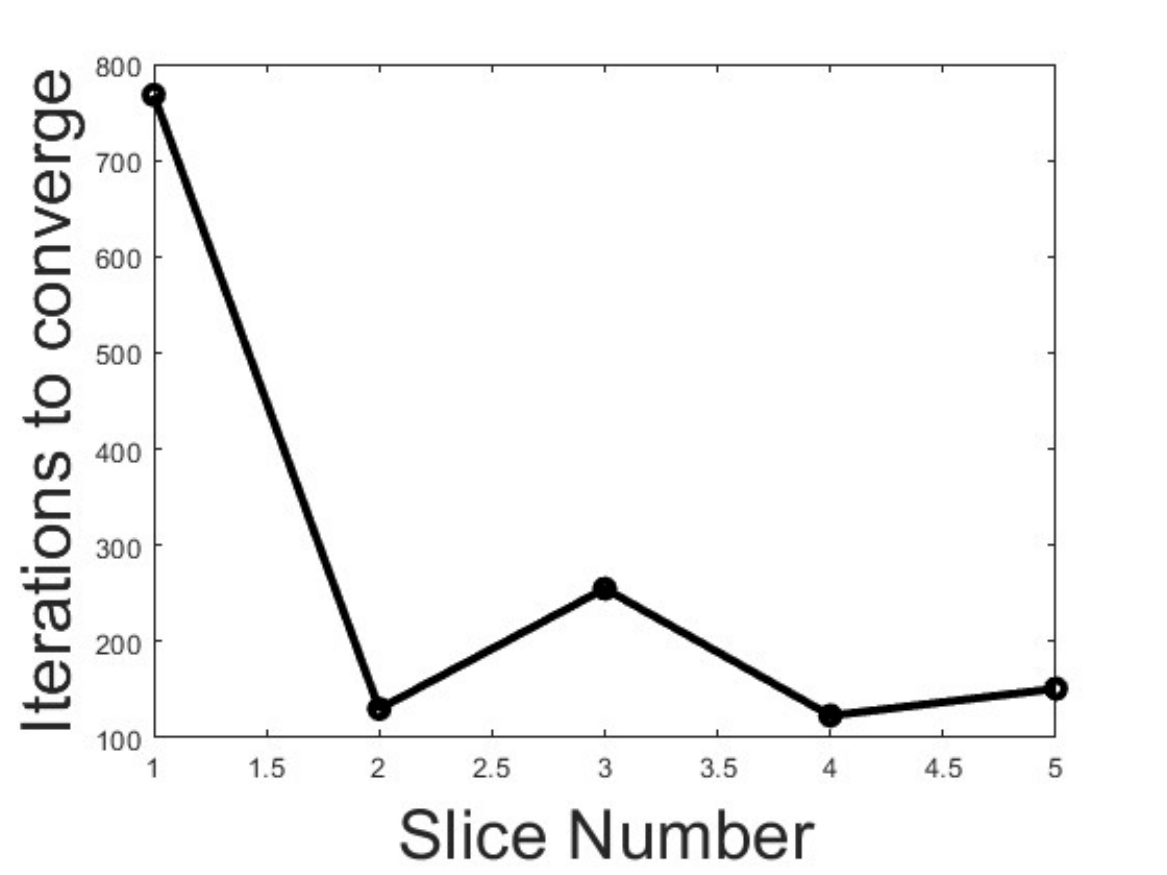}
			\caption{Faster convergence due to transfer learning.}
			\label{fig: Speed_Up}
		\end{figure}
		\begin{figure}[ht!]
			\centering
			\includegraphics[width=1.1\linewidth]{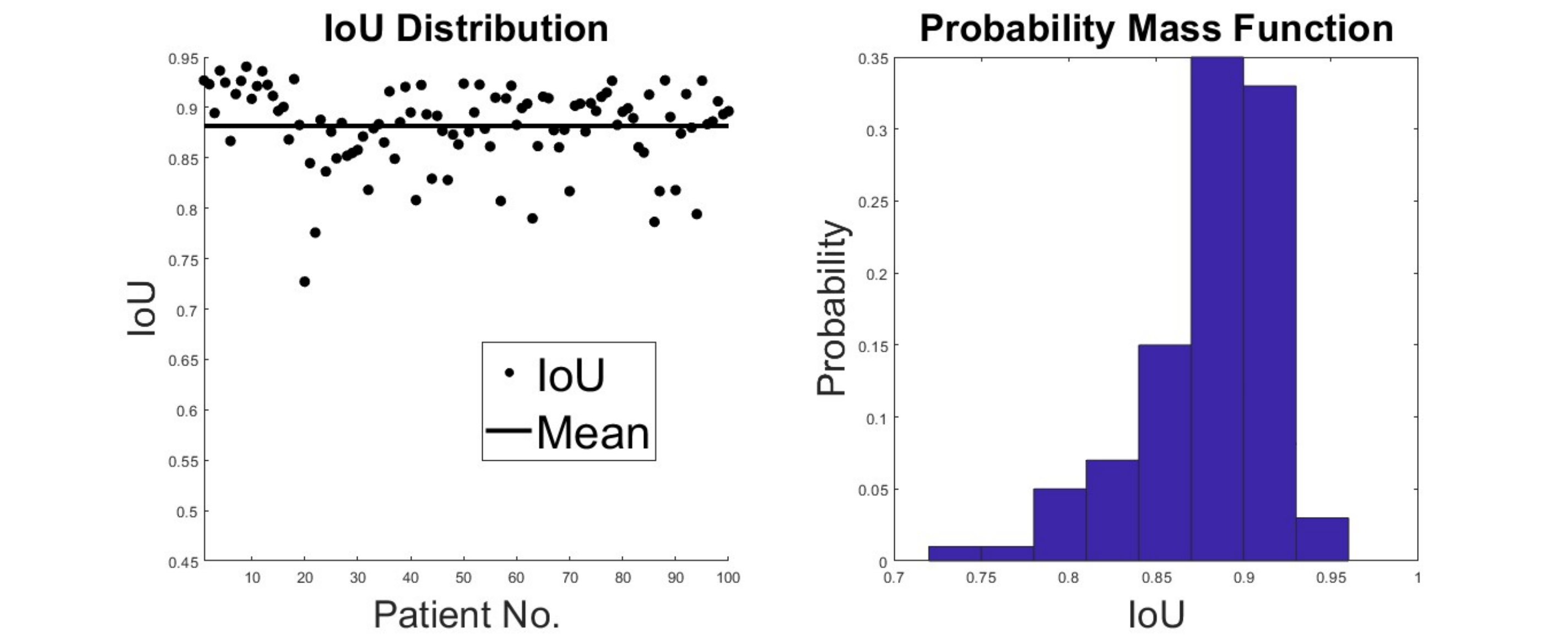}
			\caption{Performance of PICS in annotating the whole dataset consisting of 100 patients.}
			\label{fig: 3D_Seg_LV_Final}
		\end{figure}
		\begin{figure}[ht!]
			\centering
			\includegraphics[width=0.9\linewidth]{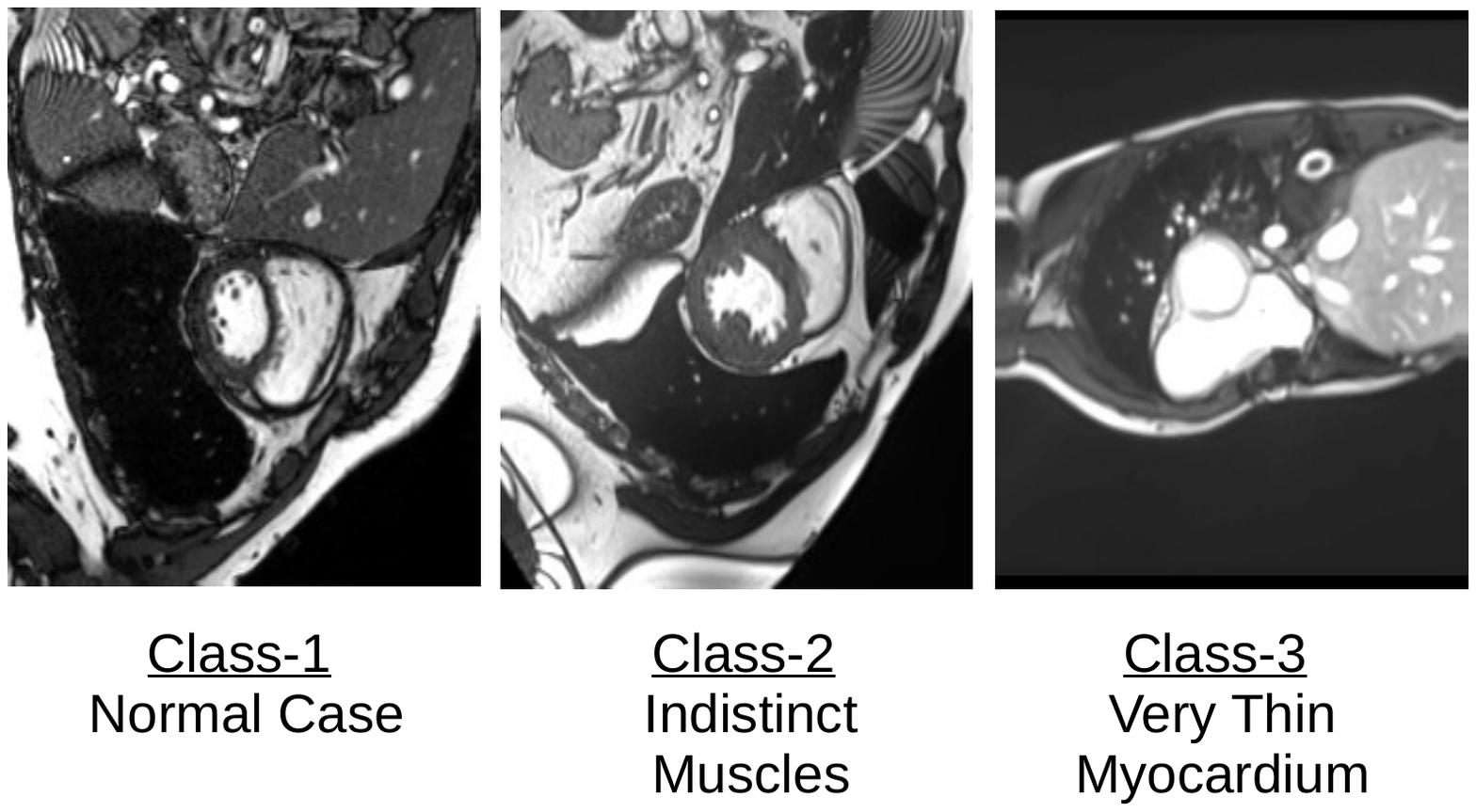}
			\caption{Hyperparameter selection for three distinct classes.}
			\label{fig: Three_Classes}
		\end{figure}
		
		\item \textbf{Limitation of PICS}
		\begin{enumerate}
			\item \textit{Inverse parameter estimation.} It may be argued that because PINNs have been successful in both forward and inverse problems and PICS is derived from PINN, it should be possible to estimate hyperparameters by minimizing a loss function with hyperparameters treated as trainable weights. However, Zapf \textit{et al.} \cite{zapf2022investigating} explain that this may not be effective. Even when the predicted segmentation is correct, all three terms of the PICS loss function, including shape regularization, region-based loss, and shape-based loss, may not be equal to zero. Depending on the image complexity, the ratio of region-based loss to other losses may be the most trustworthy at times, while the ratio of shape-based loss to others may be most trustworthy at other times. Therefore, in this work, hyperparameters were chosen through trial and error. For simpler images, like those in Figure \ref{fig: Cavity_Knots}, we can select the ratio of region-based loss to other losses as the most trustworthy and automate the hyperparameter selection process. Readers can appreciate that the proposed OPI effectively implements the suggestion by Zapf \textit{et al.} \cite{zapf2022investigating}.
			\item \textit{Topology change.} The PICS framework for 3D segmentation has a limitation regarding images that change topology during segmentation. In such cases, multiple initializations are required, but the number of initializations needed is fixed based on the topology of the first image. This limitation is demonstrated in Figure \ref{fig: Topology_Break}, where the 3D segmentation of CT scans of a hydrocephalus patient starts with one object in the first image but breaks into two parts in the fourth image, causing PICS to struggle with the increased number of parts. This issue may affect the accuracy and efficiency of the segmentation in cases where topology changes occur frequently.
			\item \textit{Less number of heart section images.} If there are only a few slices available for a specific cardiac cycle phase, such as end-diastole (ED), it could result in a significant change in the size of the left ventricle across adjacent slices, which can affect the accuracy of segmentation. Hence, having a larger number of slices available for a given phase is better for accurate segmentation of the left ventricle.
		\end{enumerate}
	\end{itemize}
	
	\begin{figure}[ht!]
		\begin{centering}
			\includegraphics[width=0.9\linewidth]{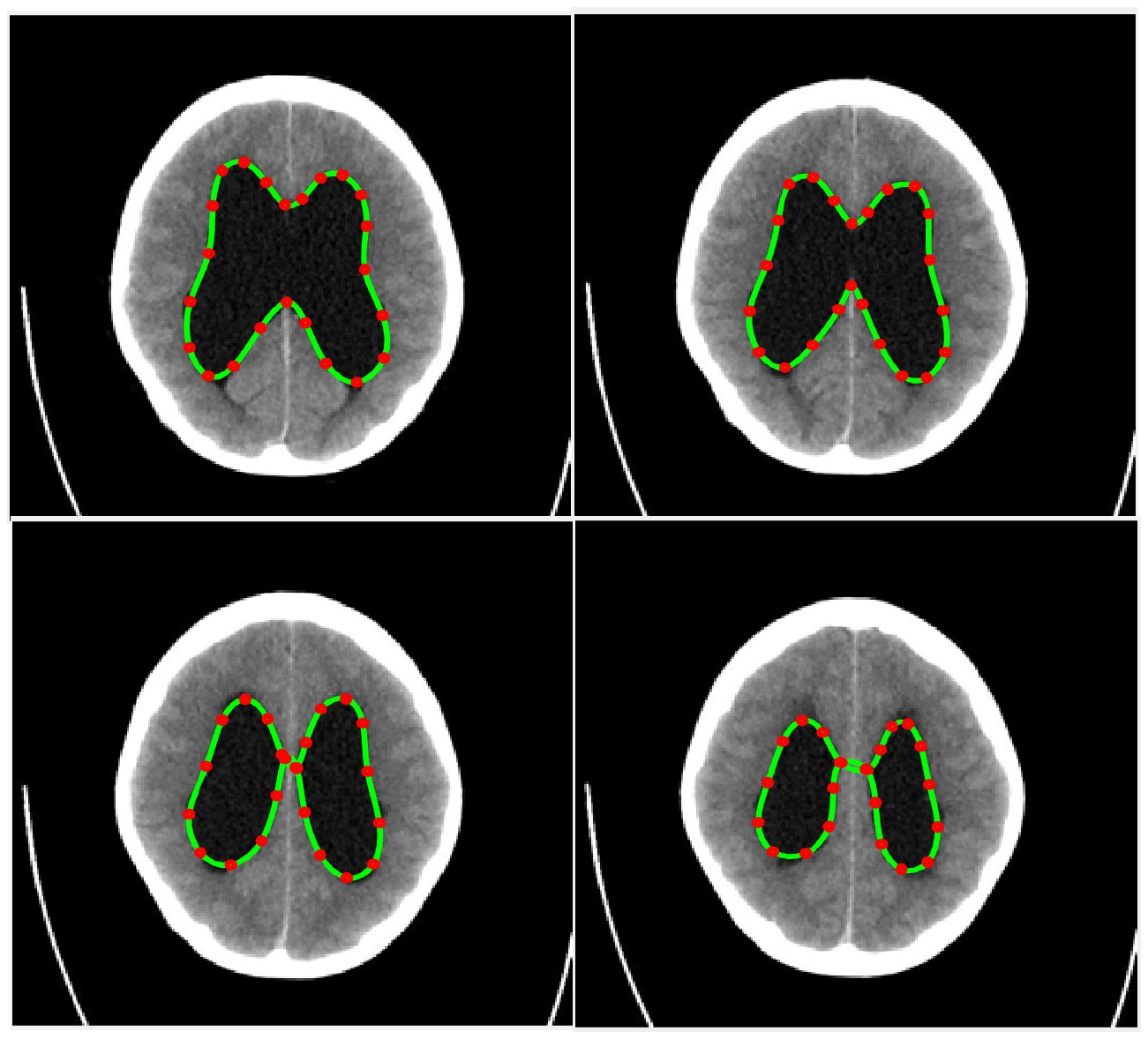}
			\par\end{centering}
		\caption{\label{fig: Topology_Break} An example of topology change: 3D segmentation of enlarged ventricles of a hydrocephalus patient.}
	\end{figure}
	\subsection{Adaptive hyperparameter tuning with OPI}
\begin{figure}[ht!]
		\begin{centering}
			\includegraphics[width=0.7\linewidth]{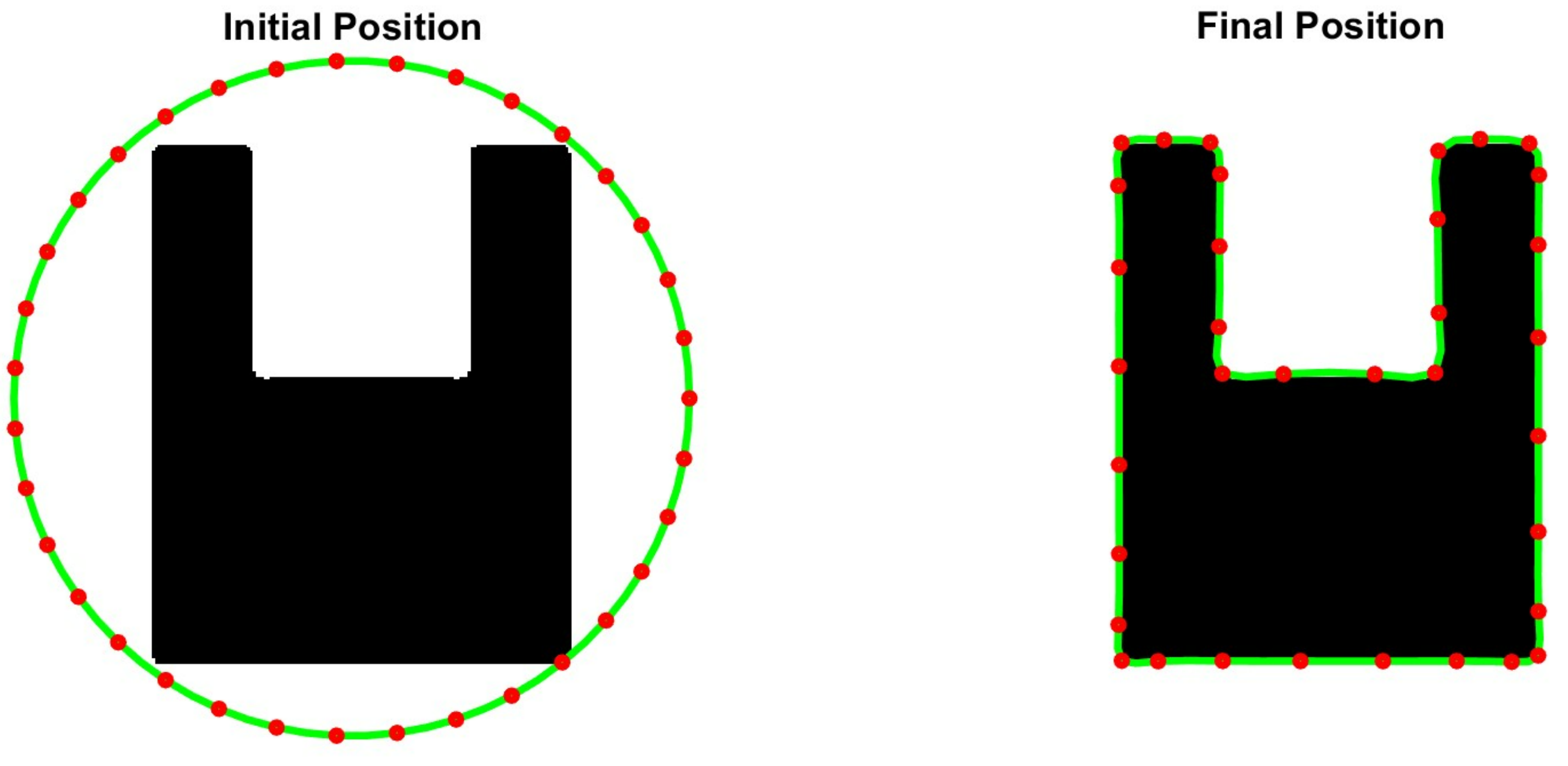}
			\par\end{centering}
		\caption{\label{fig: Cavity_Knots}Performance of PICS with adaptive hyperparameters on a u-shaped cavity.}
	\end{figure}

\begin{figure}[ht!]
		\begin{centering}
			\includegraphics[width=0.9\linewidth]{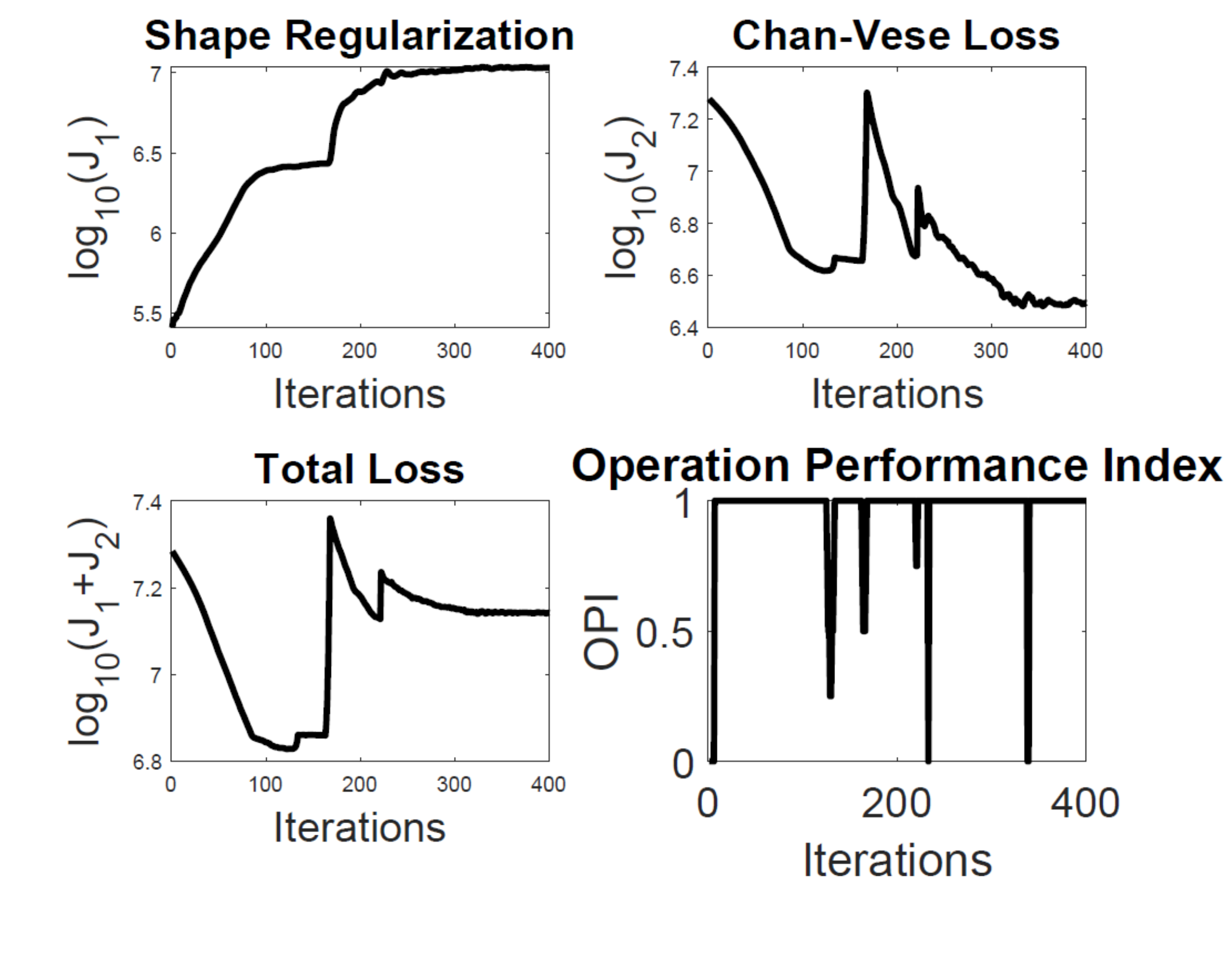}
			\par\end{centering}
		\caption{\label{fig: Cavity_OPI} OPI trend of PICS with adaptive hyperparameters for cavity test case.}
	\end{figure}

\begin{figure}[ht!]
		\begin{centering}
			\includegraphics[width=0.45\linewidth]{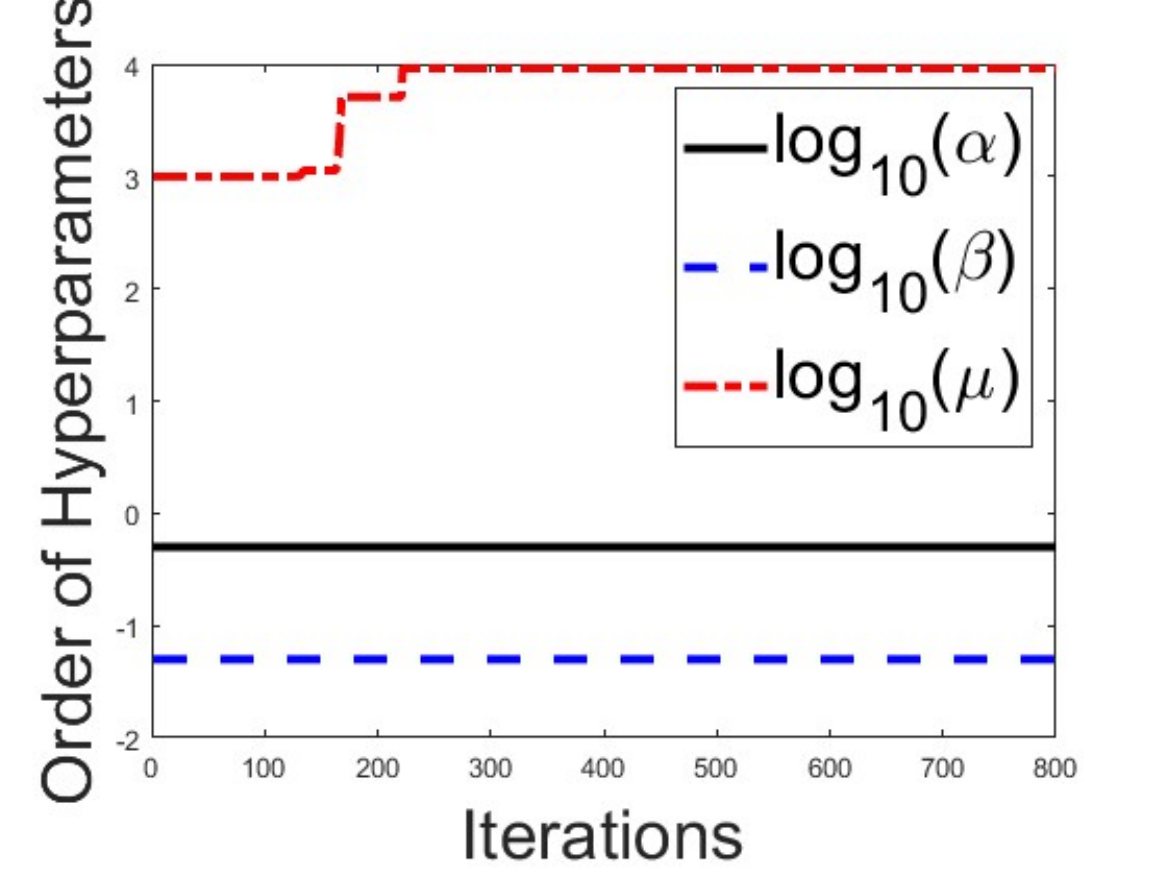}
			\par\end{centering}
		\caption{\label{fig: Cavity_Adaptive} PICS hyperparameters tuning for cavity test case.}
	\end{figure}

\begin{figure}[ht!]
		\centering
		\includegraphics[width=0.9\linewidth]{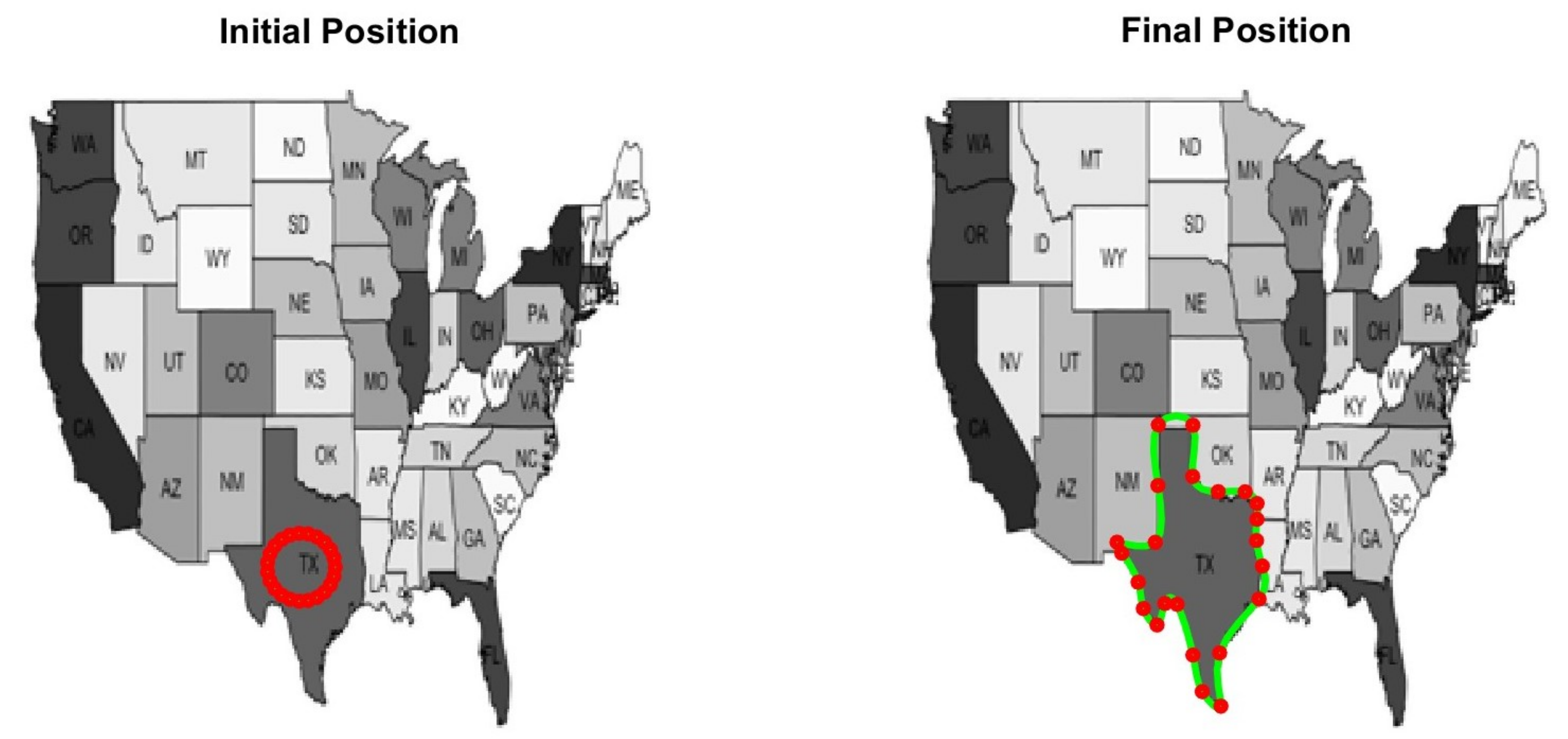}
		\caption{Segmentation of Texas state from the USA map. Left: Initial weights, Right: Optimized weights.}
		\label{fig: 2D_Seg_Texas}
	\end{figure}

\begin{figure}[ht!]
		\begin{centering}
			\includegraphics[width=0.45\linewidth]{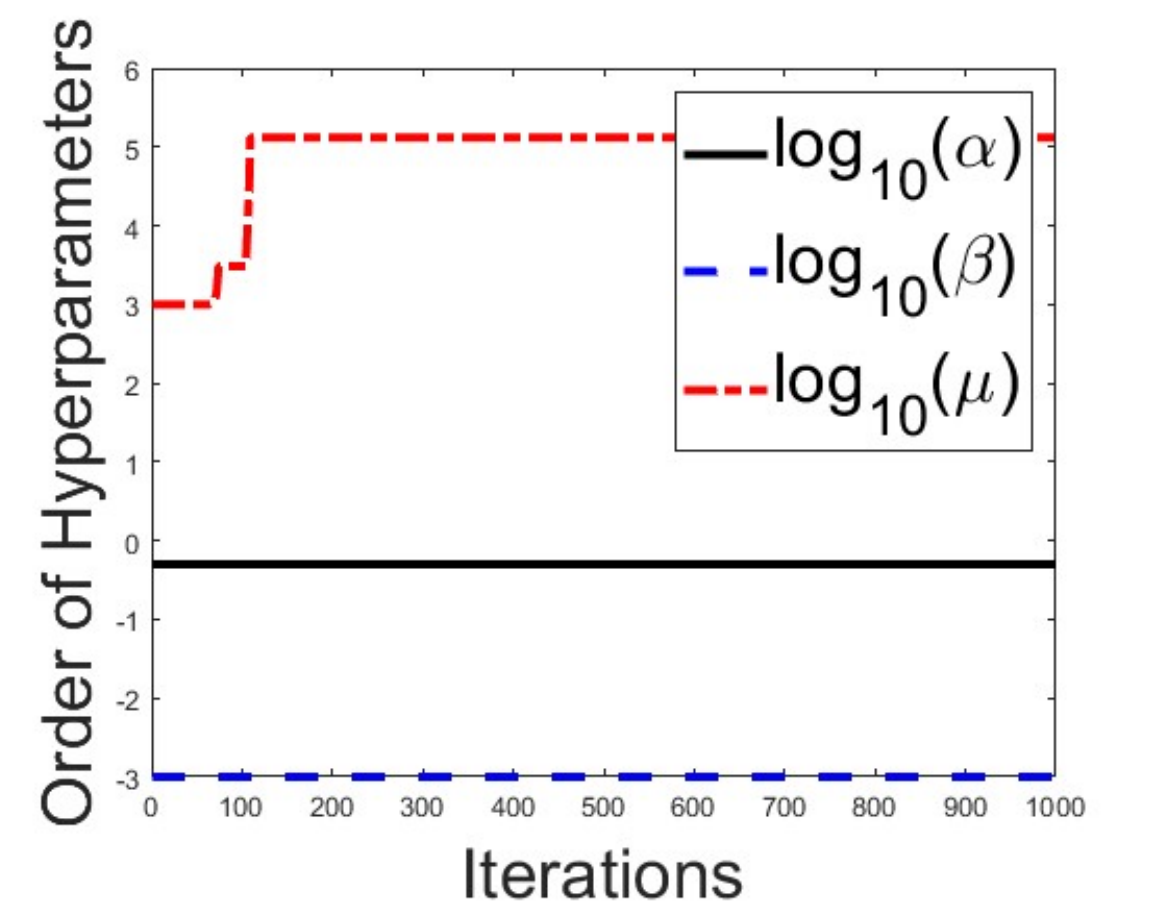}
			\par\end{centering}
		\caption{\label{fig: Texas_Adaptive} PICS hyperparameters tuning for the USA map case.}
	\end{figure}

	\begin{itemize}
		\item \textit{A global minimum and a local minimum.} Figure \ref{Fig_bad_minima} displays two instances of bad minima. The first example shows a shrunken snake due to a high value of the bending coefficient. In contrast, the second shows the snake getting trapped in a local minimum because the increase in loss value caused by extension is greater than the drop in loss value due to the Chan-Vese loss. The loss history and OPI trend for each case are shown in Figures \ref{fig: OPI_LM_01} and \ref{fig: OPI_LM_02} respectively. When examining the total loss history alone, it is difficult to determine whether the optimization is progressing correctly. However, looking at the OPI trend, we can see that its value is zero for the first case and fluctuates wildly between 0 and 1 for the second. Therefore, we can rely on the OPI to determine that PICS is stuck in a local minimum and requires adjustments to its hyperparameters to overcome this issue.
		\begin{figure}[ht!]
			\centering
			\begin{minipage}{0.35\textwidth}
				\centering
				\includegraphics[width=\linewidth]{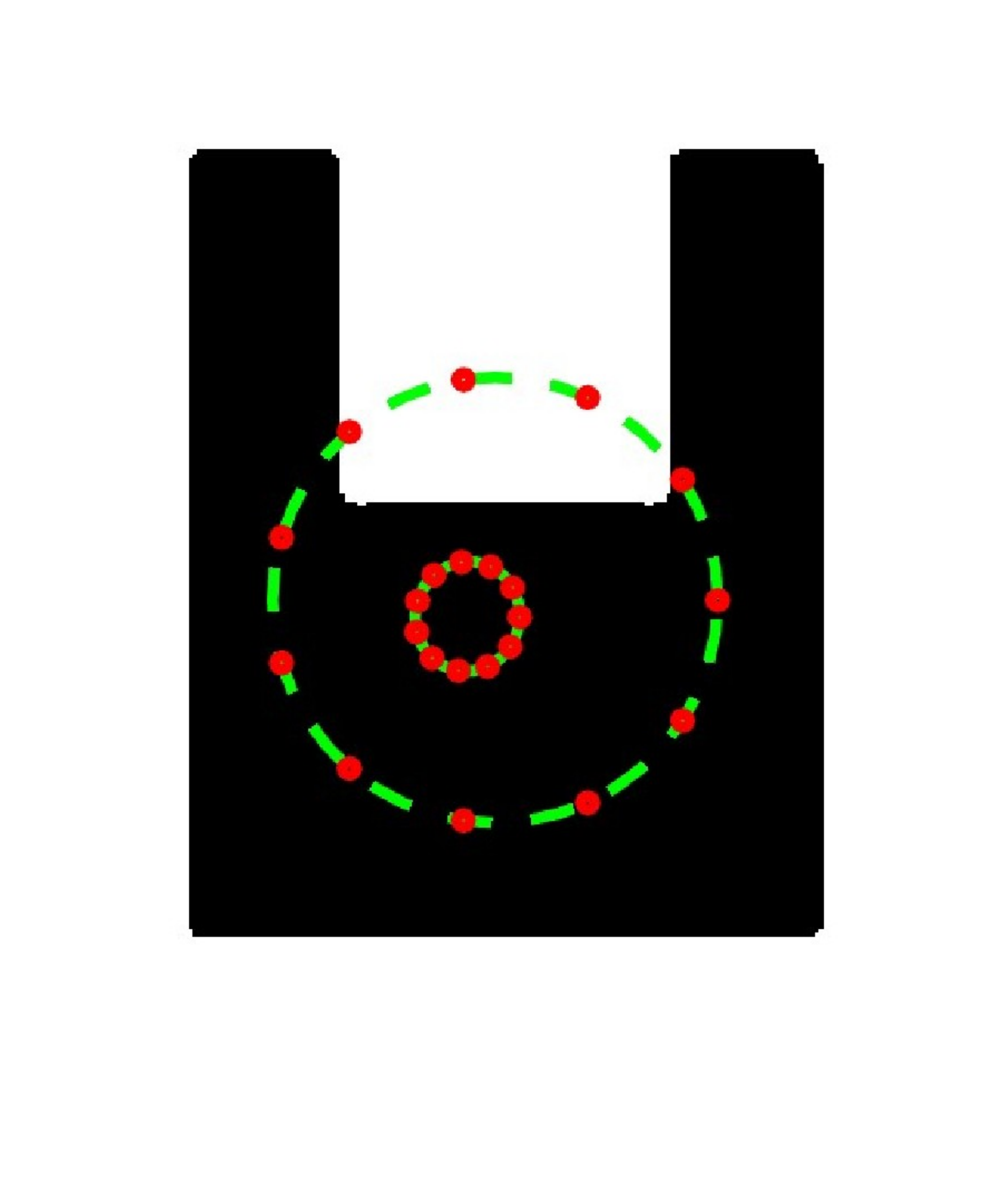}
			\end{minipage}
			\begin{minipage}{0.35\textwidth}
				\centering
				\includegraphics[width=\linewidth]{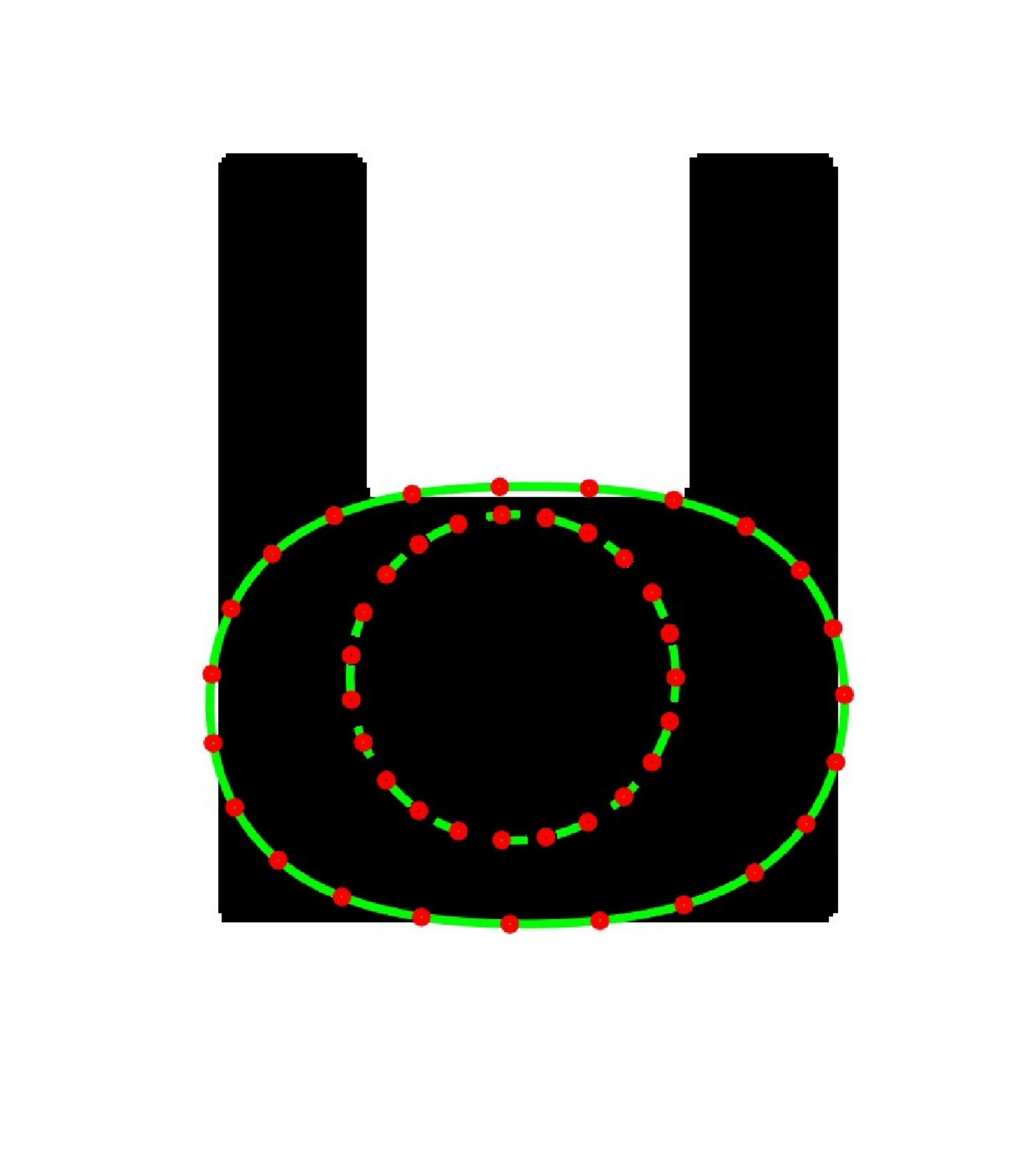}
			\end{minipage}
			\caption{\label{Fig_bad_minima}Two examples of bad minima. The initial snake is shown in dashed lines, and the final snake is shown in continuous lines.}
		\end{figure}
		\begin{figure}[ht!]
			\centering
			\includegraphics[width=0.7\linewidth]{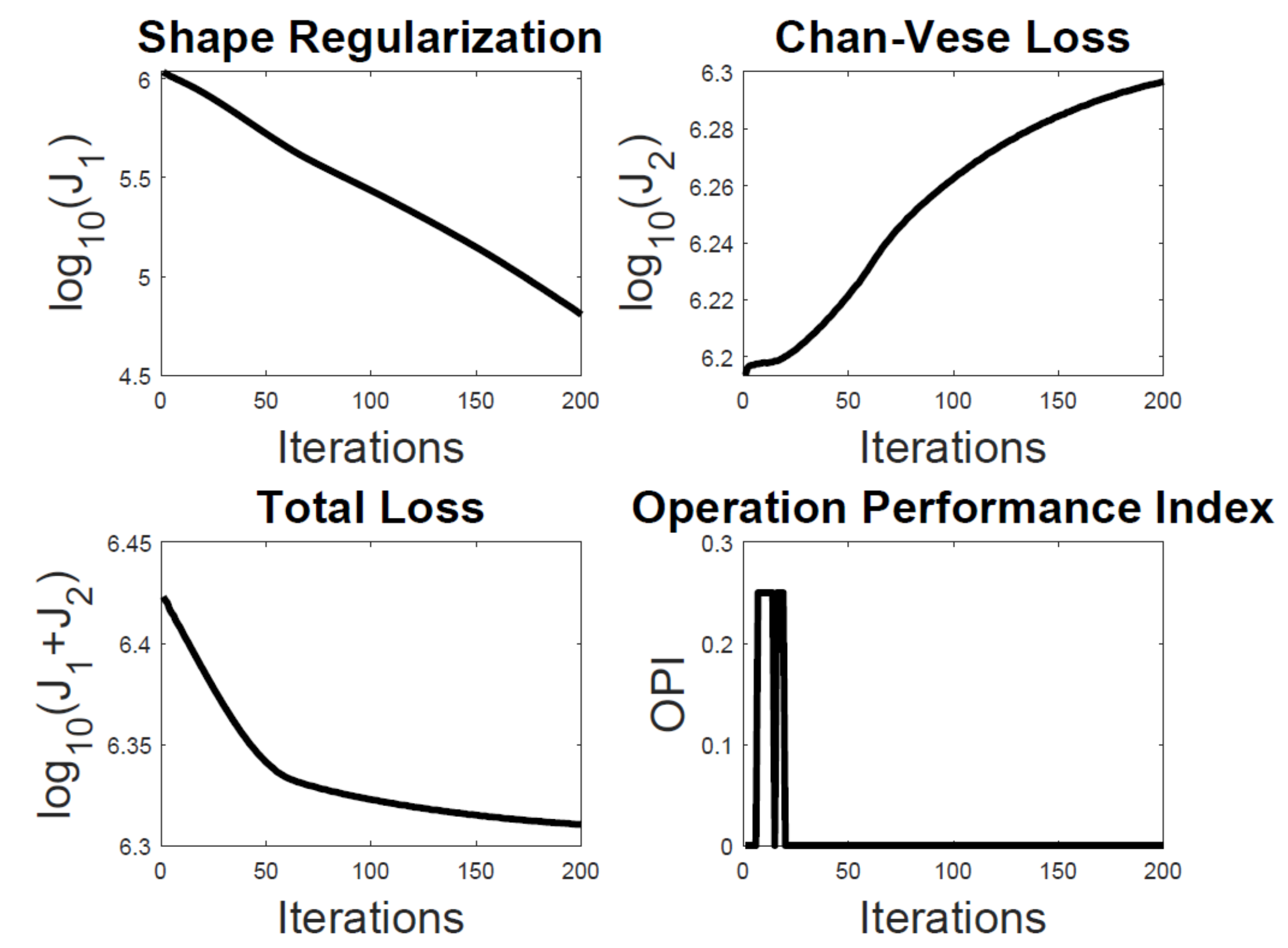}
			\caption{OPI trend (low value throughout) for LHS case of Fig.\ref{Fig_bad_minima}. The snake shrinks at all the steps and will ultimately shrink to a point.}
			\label{fig: OPI_LM_01}
		\end{figure}
		\begin{figure}[ht!]
			\centering
			\includegraphics[width=0.7\linewidth]{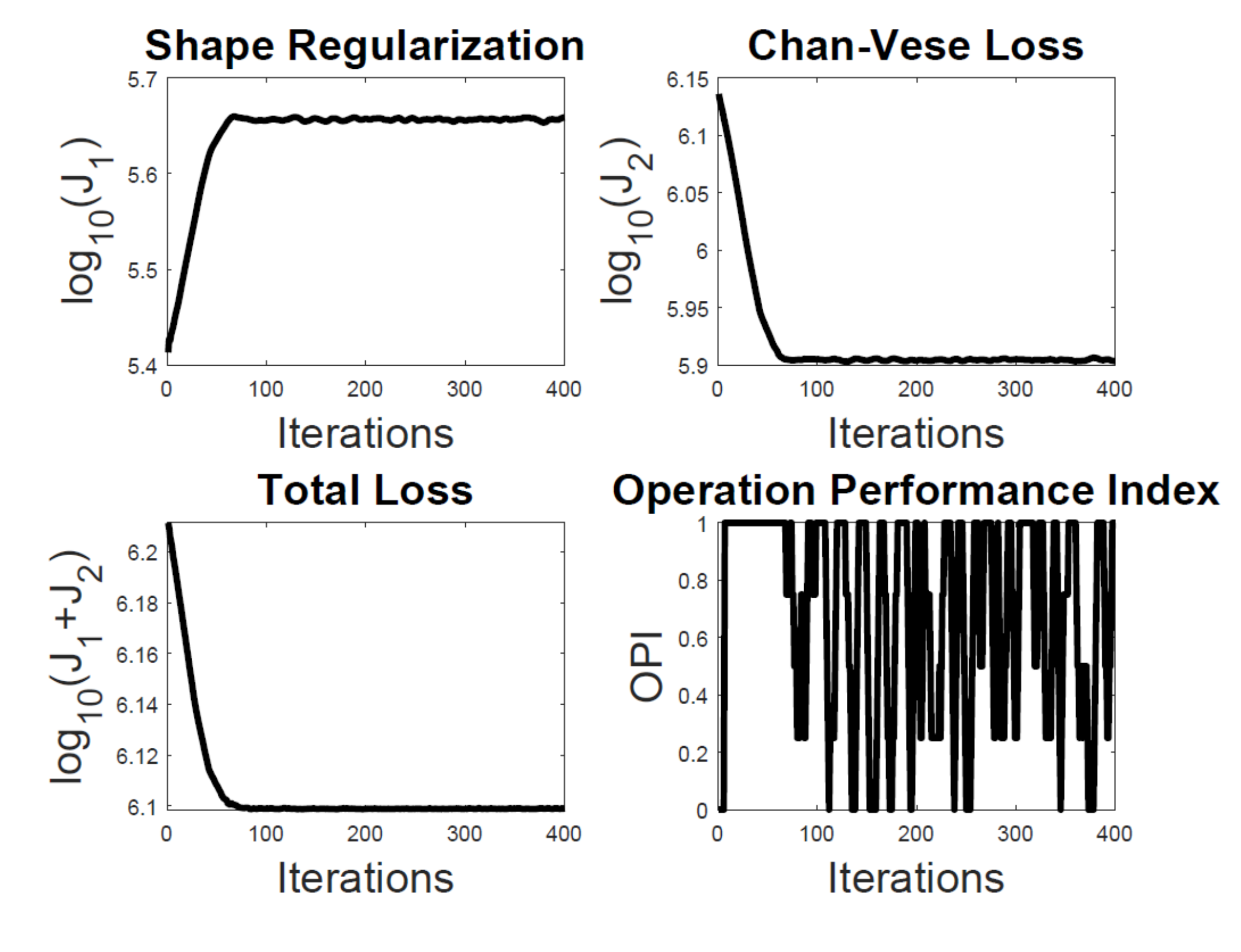}
			\caption{OPI trend (oscillating between 0 and 1) for RHS case of Fig.\ref{Fig_bad_minima}. The snake gets stuck in a local minimum after about 100 iterations.}
			\label{fig: OPI_LM_02}
		\end{figure}
		\item \textit{Adaptive hyperparameter tuning.} Figure \ref{fig: Cavity_Knots} demonstrates that PICS is capable of accurately capturing concave regions. Figures \ref{fig: Cavity_OPI} and \ref{fig: Cavity_Adaptive} provide additional information on this particular case, including the OPI score, loss history, and adaptive tuning of hyperparameters. Similar results are shown in the figs.\ref{fig: 2D_Seg_Texas} and \ref{fig: Texas_Adaptive}. In this case, we segment Texas state borders from the USA map. These figures show that hyperparameters are adjusted as needed when the OPI drops and a high value of OPI ensures that the optimization is always progressing in the correct direction.
		
	\end{itemize}
	
	\section{Conclusions and Future Work} \label{Conclusion}
	In this paper, we introduced PICS-- an interpretable, physics-informed algorithm for rapid image segmentation in the absence of labeled data. PICS is a novel algorithm that combines the traditional active contour model called snake with the physics-informed neural networks (PINNs). PICS inherits the unique qualities of its parent algorithms (snakes and PINNs), making it intuitive, mesh-free, and respecting the inherent physics of the traditional energy-based loss functions. The use of cubic splines over deep neural network as basis function and the treatment of spline control knots as design variables further increase its interpretability. We demonstrate that PICS is the first work to minimize the Chan-Vese loss in the snake framework and allows for easy integration of medical domain expertise via prior shape-based loss terms. PICS draws a parallel between 3D segmentation and the solution of an unsteady PDE. The results obtained on the ACDC dataset show that it allows PICS to exploit transfer learning and achieve fast segmentation. However, PICS faces challenges in inverse parameter estimation and topology changes during 3D segmentation. To address some of these challenges, we propose a new evaluation metric called Optimization Performance Index (OPI), which allows for adaptive hyperparameter tuning. Overall, PICS demonstrates its effectiveness as an alternative to deep learning-based segmentation models in the absence of labeled data. It offers a promising solution for medical image segmentation with its speed, computational efficiency, and human-centered approach, which can significantly reduce the time and cost associated with obtaining labeled training data.

	\section*{Acknowledgements}
	This work was supported by the Robert Bosch Centre for Data Science and Artificial Intelligence, Indian Institute of Technology- Madras, Chennai (Project no. SB21221651MERBCX008832).
	
	\bibliographystyle{unsrt}  
	\bibliography{references}

\end{document}